\documentclass{article} 
\usepackage{colm2024_conference}

\usepackage{booktabs}
\usepackage{graphicx}
\usepackage{enumitem}
\usepackage{wrapfig}
\usepackage{longtable}
\usepackage{algorithm}
\usepackage{algpseudocode}

\usepackage{amsmath,amsfonts,bm}

\def\eqref#1{equation~\ref{#1}}

\def\1{\bm{1}}

\DeclareMathAlphabet{\mathsfit}{\encodingdefault}{\sfdefault}{m}{sl}
\SetMathAlphabet{\mathsfit}{bold}{\encodingdefault}{\sfdefault}{bx}{n}

\hypersetup{
  hypertexnames=false,
  pdftitle={Qwen-CUA: Native Computer Use for (almost) Everything},
  pdfauthor={Qwen Team and Xlang Lab},
  pdfsubject={Technical Report}
}

\makeatletter
\renewenvironment{abstract}
  {\vskip.075in\centerline{\large\bf Abstract}\vspace{0.5ex}\begin{quote}}
  {\par\end{quote}\vskip 1ex}
\makeatother

\renewcommand{\hflink}{}
\renewcommand{\ghlink}{https://github.com/xlang-ai/Qwen-CUA}

\newcommand{\qwencuatitleicon}{\makebox[0pt][r]{\raisebox{-0.72em}{\includegraphics[height=2.50em,trim=140 90 90 90,clip]{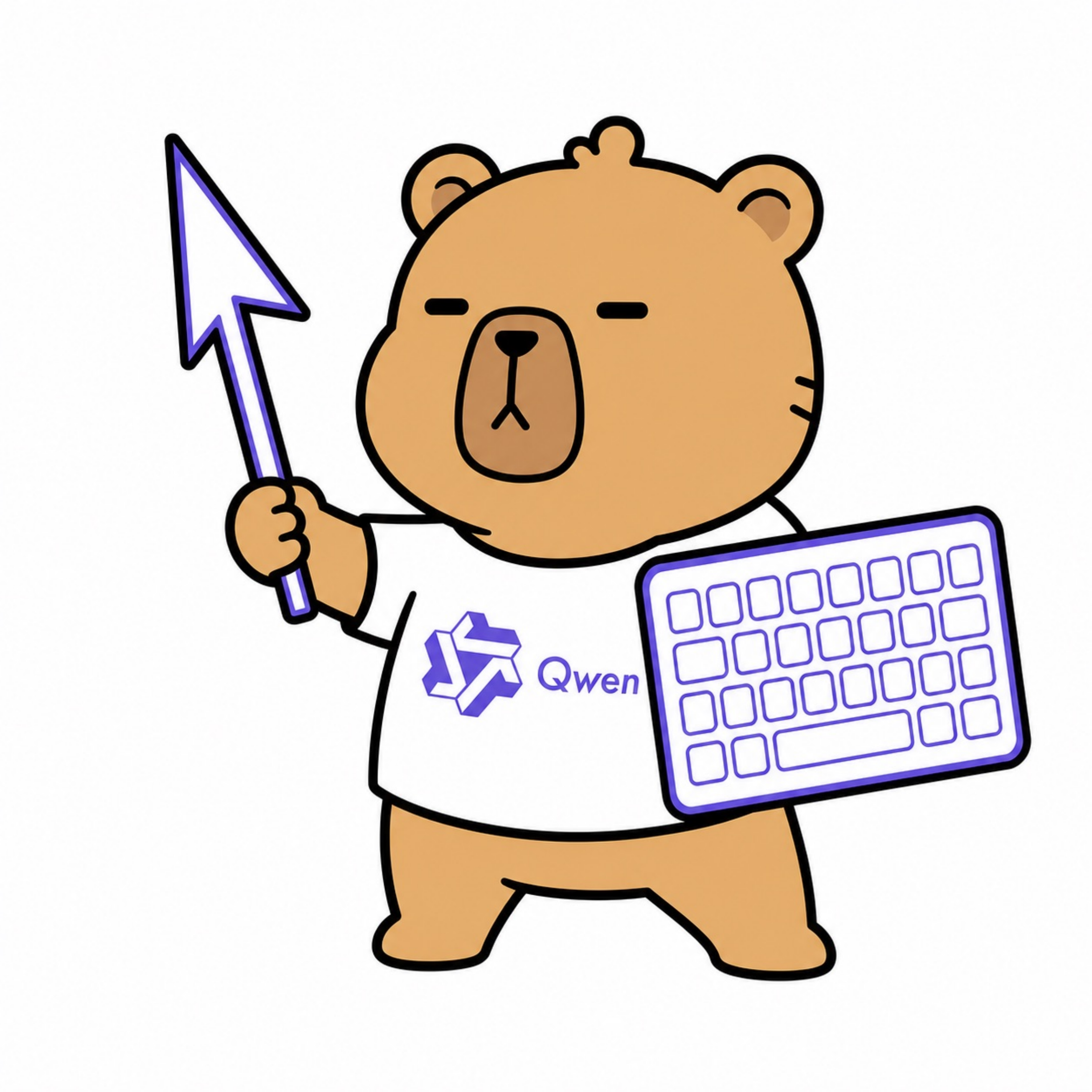}}\hspace{0.35em}}}
\title{\qwencuatitleicon Qwen-CUA: Native Computer Use for (almost) Everything}

\author{\textbf{Qwen Team \& Xlang Lab}}

\newcommand{\corecontributors}{%
Shuai Bai,
Tianyi Bai\textsuperscript{*},
Sicheng Fan,
Chang Gao,
Jian Guan,
Feng Hu,
Mianqiu Huang,
Xingyang Huang,
Yizhen Jiang,
Yuheng Jing,
Dehui Kong,
Ning Li,
Dayiheng Liu,
Shixuan Liu\textsuperscript{*},
Zheng Liu,
Dunjie Lu,
Que Shen,
Bowen Wang\textsuperscript{*},
Junli Wang\textsuperscript{*},
Chencan Wu,
Rui Xie,
Tianbao Xie\textsuperscript{*},
Zhihui Xie,
Haiyang Xu,
An Yang,
Tao Yu,
Wenzhen Yuan,
Xi Zhang,
Zhenru Zhang,
Mingkang Zhu,
Zhaoqing Zhu%
}

\newcommand{\contributors}{%
Yizhong Cao,
Kai Dang,
Binyuan Hui\textsuperscript{*},
Kaixin Li\textsuperscript{*},
Junyang Lin\textsuperscript{*},
Haiquan Wang,
Zekun Wang,
Yiheng Xu\textsuperscript{*},
Fan Yan,
Mengqi Yuan,
Danyang Zhang\textsuperscript{*},
Jiajun Zhang,
Zhipeng Zhang\textsuperscript{*},
Fan Zhou\textsuperscript{*}, 
Fan Zhou
}

\begin{document}

\maketitle

\begin{abstract}
Native computer use offers a general route to agents that can operate almost
any software through the same interface available to people. Realizing this
promise, however, requires more than visual grounding: an agent must sustain
long-horizon state, acquire large amounts of costly interactive experience, and
learn from outcomes that are sparse but reliably verifiable. We introduce
Qwen-CUA, a native computer-use agent with a 397B-A17B Qwen
mixture-of-experts backbone. It observes only screenshots and acts through keyboard and mouse
events, without DOM trees, accessibility metadata, or task-specific APIs. Its
agent scaffold expands the active visual history to 20 screenshots and folds
older screenshots in fixed-size blocks, retaining recent visual evidence while
preserving reusable prompt prefixes. To scale training, we build a cloud rollout
fleet with access to nearly 100{,}000 vCPUs and tens of thousands of concurrent
environments, construct approximately 40{,}000 verifiable tasks, and collect
personalized long-horizon workflows in everyday and
professional software. We optimize complete trajectories with verifiable
rewards and long-horizon trajectory slicing, and develop the model through
iterative training runs that use each resulting policy to refresh the
supervised data mixture and calibrate the reinforcement-learning task
distribution. Across eight computer-use benchmarks, Qwen-CUA
consistently outperforms Qwen3.7 and remains competitive with leading
proprietary systems, reaching 86.2 on OSWorld-Verified and 18.5 / 48.4 binary /
partial completion on OSWorld 2.0. Scaling the same recipe to a model with over
one trillion total parameters yields Qwen-CUA-Max, which further improves
these scores to 87.6 and 21.2 / 53.3. Qwen-CUA also reduces attack success on
RedTeamCUA from 36.6 to 16.4 relative to Qwen3.7. Efficiency analyses, an internal browser
deployment, and experiments combining native interaction with Bash further
characterize its practical behavior. These results show that native computer
use can serve as a broadly capable foundation, while scalable verifiable
interaction and hybrid tool use are key to making it effective in practice.
\end{abstract}

\vspace{-0.5em}
\begin{center}
\includegraphics[width=0.97\linewidth]{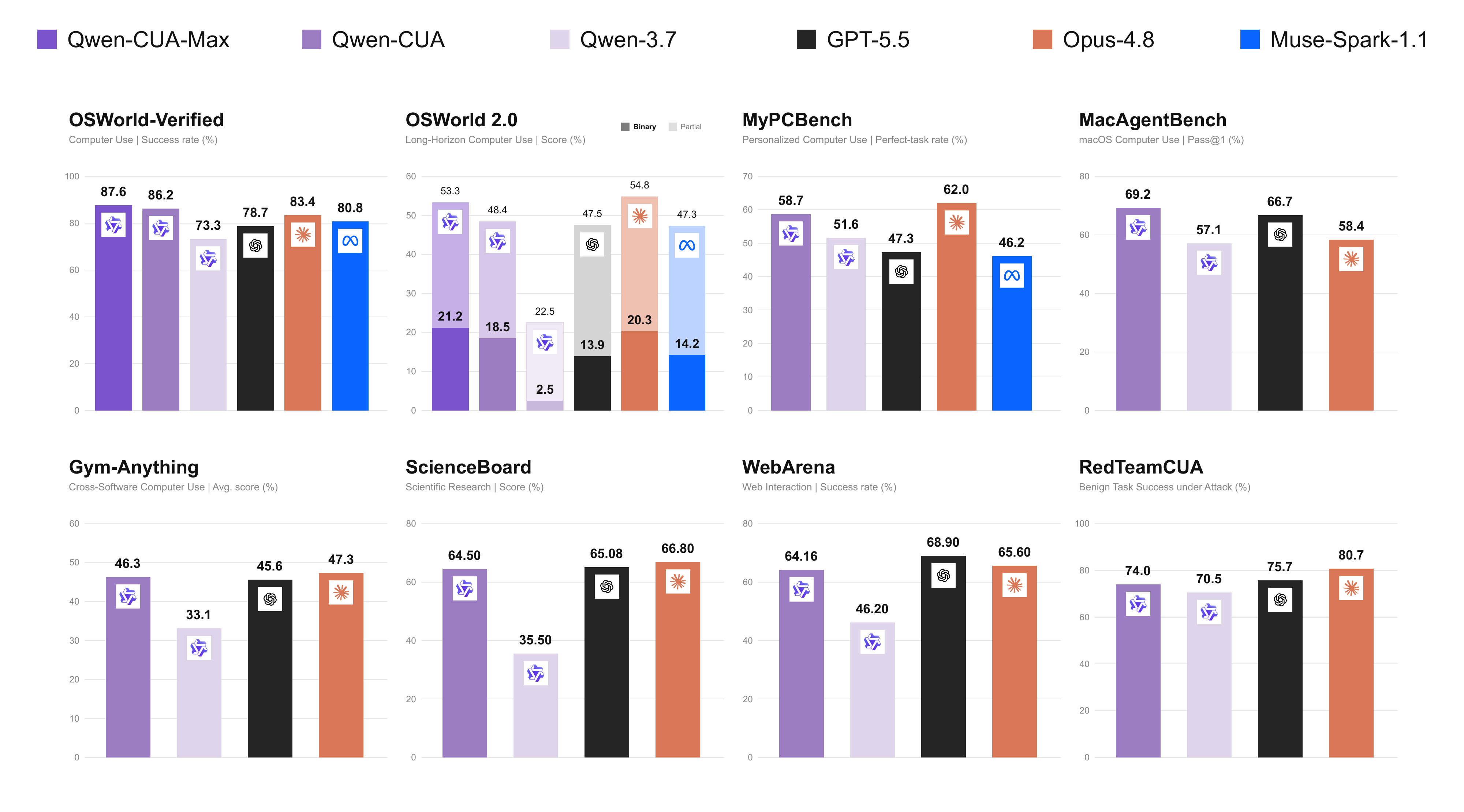}

\vspace{-0.25em}
{\small\textbf{Figure 1:} Main results across eight computer-use
benchmarks. OSWorld 2.0 reports binary completion (dark) and partial completion
(light), while RedTeamCUA reports benign task success under attack. All other
panels show a single score. Higher is better.}
\end{center}

\newpage
\section{Introduction}

Agents act on the digital world through three broad interfaces: code, APIs, and
graphical interfaces designed for people. Foundation models have become highly
capable at the first two, writing programs and composing software tools. Yet
much digital work lies beyond them: desktop applications, legacy systems,
dynamic websites, professional tools, and personalized workflows often expose
functionality only visually. Native computer use closes this gap. By perceiving
pixels and producing keyboard-and-mouse events, an agent can operate the same
software as a person without dedicated integrations, unlocking workflows
previously accessible mainly through direct human interaction.

This human-facing interface poses distinct learning challenges. GUI state is
partially observed and machine-unreadable; actions require pixel grounding,
errors compound over long workflows, and evidence spans screenshot history.
Benchmarks now cover open-ended web and operating-system tasks
~\citep{zhou2023webarena,xie2024osworld}, but each rollout consumes a stateful
environment and real interaction time, while reliable supervision often
arrives only from the final state.
Agents must also generalize to personalized desktops, professional software,
simulated users, and adversarial on-screen content
~\citep{jang2026mypcbench,sun2025scienceboard,
aggarwal2026gymanything,liao2025redteamcua}. Progress therefore requires scaling
environments, verifiable tasks, rollout throughput, and learning from long
multimodal trajectories.

We present \textbf{Qwen-CUA}, a Qwen model trained end to end for native computer
use. It receives only screenshots and emits keyboard-and-mouse actions, without
DOM trees, accessibility metadata, shell access, or task-specific APIs. This
minimal interface transfers unchanged across browsers and desktop applications.
For long-horizon operation, Qwen-CUA retains 20 active screenshots and folds
older screenshots in blocks of ten, bounding visual context while keeping the
prompt prefix stable for improved KV-cache reuse.

To produce verifiable interaction at scale, we build an Alibaba Cloud ECS
rollout fleet with access to nearly 100{,}000 vCPUs and tens of thousands of
concurrent environments. The environment pool combines controllable mock web
services with everyday, long-tail, and professional desktop applications. We
construct approximately 40{,}000 verifiable tasks spanning
environment operation, simulated-user interaction, and long-horizon workflows
constructed through verifiable phase-state chaining. Human trajectories from
personalized desktops and specialized software provide complementary
supervision, augmented with action-grounded step-level reasoning.

Qwen-CUA is optimized with executable outcome rewards, building on the scalable
environment and task design of CUA-Gym~\citep{wang2026cuagym}. Soft Adaptive
Policy Optimization (SAPO)~\citep{gao2025sapo} stabilizes updates over long
reasoning-and-action sequences, while trajectory slicing preserves episode-level
rewards across image-heavy histories. Training proceeds in iterations: each
resulting model reveals unresolved SFT queries and weak domains, guiding teacher
rerolls, new human trajectories, targeted data collection, and RL task
calibration for the next run.

Under a native keyboard-and-mouse computer-use protocol, Qwen-CUA outperforms
Qwen3.7 on the task
metric of all eight benchmarks. It reaches 86.2 on OSWorld-Verified, compared
with 73.3 for Qwen3.7, 78.7 for GPT-5.5, and 83.4 for Claude Opus 4.8, and
improves
OSWorld 2.0 binary / partial completion from 2.5 / 22.5 to 18.5 / 48.4. Scaling
the 397B-A17B model to over one trillion total parameters yields
\textbf{Qwen-CUA-Max}, further increasing these results to 87.6 and 21.2 / 53.3.
Qwen-CUA also reduces RedTeamCUA attack success from 36.6 to 16.4 relative to
Qwen3.7 while improving task success.

Efficiency sweeps show that these gains are not explained simply by more
verbose reasoning. An internal Chrome deployment demonstrates naturally
occurring browser workflows with user confirmation for consequential actions,
while MyPCBench experiments show that combining native interaction with Bash
can substantially shorten trajectories. Together, these studies motivate
native computer use as a general visual foundation that can be combined with
specialized tools for efficient execution.

Our main contributions are:
\begin{itemize}[leftmargin=1.4em,itemsep=0.15em,topsep=0.25em]
    \item A screenshot-only Qwen-CUA model with native keyboard-and-mouse
    control and efficient long-horizon visual context management, which
    outperforms Qwen3.7 on all eight computer-use benchmarks and reaches 86.2
    on OSWorld-Verified.
    \item Scaled verifiable experience from nearly 100{,}000 vCPUs, 40{,}000
    verifiable tasks, diverse controllable environments, and personalized
    expert workflows.
    \item A comprehensive evaluation of Qwen-CUA across eight benchmarks, plus
    analyses showing that its gains are not explained by more verbose
    reasoning, that RedTeamCUA attack success drops from 36.6 to 16.4, and that
    pairing native interaction with Bash shortens trajectories; scaling the
    same recipe to the trillion-parameter Qwen-CUA-Max further lifts
    OSWorld-Verified and OSWorld 2.0 to 87.6 and 21.2 / 53.3.
\end{itemize}

\section{Agent Implementation}

Following recent screenshot-based computer-use agent
systems~\citep{xie2024osworld,xu2024aguvis,qin2025uitars,
wang2025opencua}, Qwen-CUA combines a minimal
native computer-use interface with expanded visual-history capacity and
lightweight context management for long-horizon interaction.

\subsection{Native Computer-Use Interface}
\setcounter{figure}{1}
\begin{wrapfigure}[11]{r}{0.28\linewidth}
  \vspace{-1.3em}
  \centering
  \includegraphics[width=\linewidth]{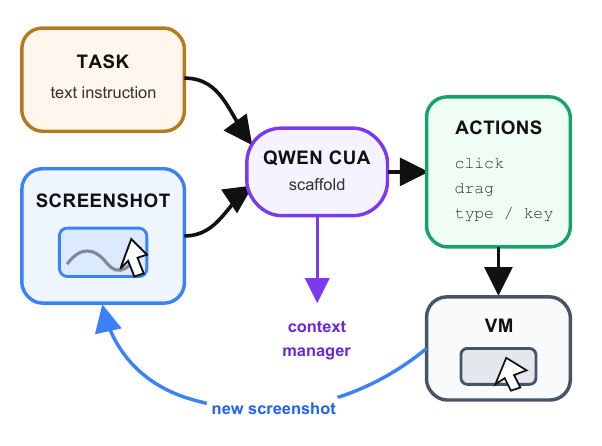}
  \refstepcounter{figure}
  \label{fig:native-computer-use-scaffold}
  \vspace{-0.15em}

  \begin{minipage}{\linewidth}
  \centering
  \footnotesize\textbf{Figure \thefigure:} Native computer-use loop.
  \end{minipage}
  \vspace{-0.8em}
\end{wrapfigure}
Qwen-CUA is implemented around a native computer-use interface. At each step, the
agent observes only a screenshot of the current desktop state and emits an action
from a keyboard-and-mouse action space. This interface deliberately avoids
task-specific APIs, accessibility trees, DOM access, or application-level
shortcuts that would expose hidden state beyond the pixels on screen. Instead,
the model controls the computer through native input events. This design makes
the agent interface close to how a human uses a desktop computer: perception is
visual, and execution is grounded in keyboard and mouse operations
(Figure~\ref{fig:native-computer-use-scaffold}). The full action schema is
listed in Appendix~\ref{app:native-action-space}.

\subsection{Long-Horizon Context Management}
\label{sec:long-horizon-context}
Long-horizon computer use produces image-heavy histories whose visual cost grows
with every interaction. Keeping all screenshots eventually exceeds the practical
context budget, while a conventional sliding window may discard the earlier state
that explains subsequent reasoning and actions.

\begin{figure}[H]
  \centering
  \begin{minipage}[t]{0.39\linewidth}
    \centering
    \includegraphics[width=\linewidth]{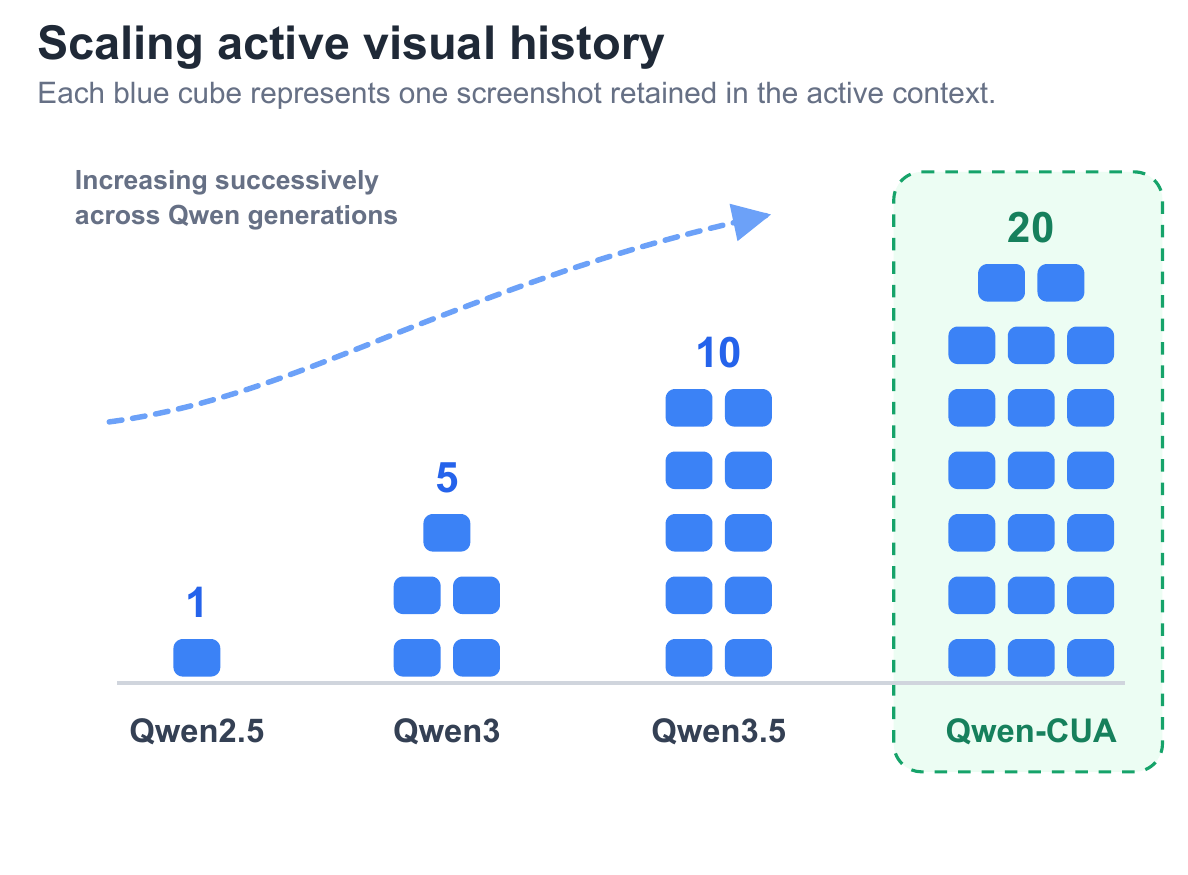}

    \vspace{-0.35em}
    \small (a)
  \end{minipage}\hspace{0.015\linewidth}
  \begin{minipage}[t]{0.46\linewidth}
    \centering
    \includegraphics[width=\linewidth]{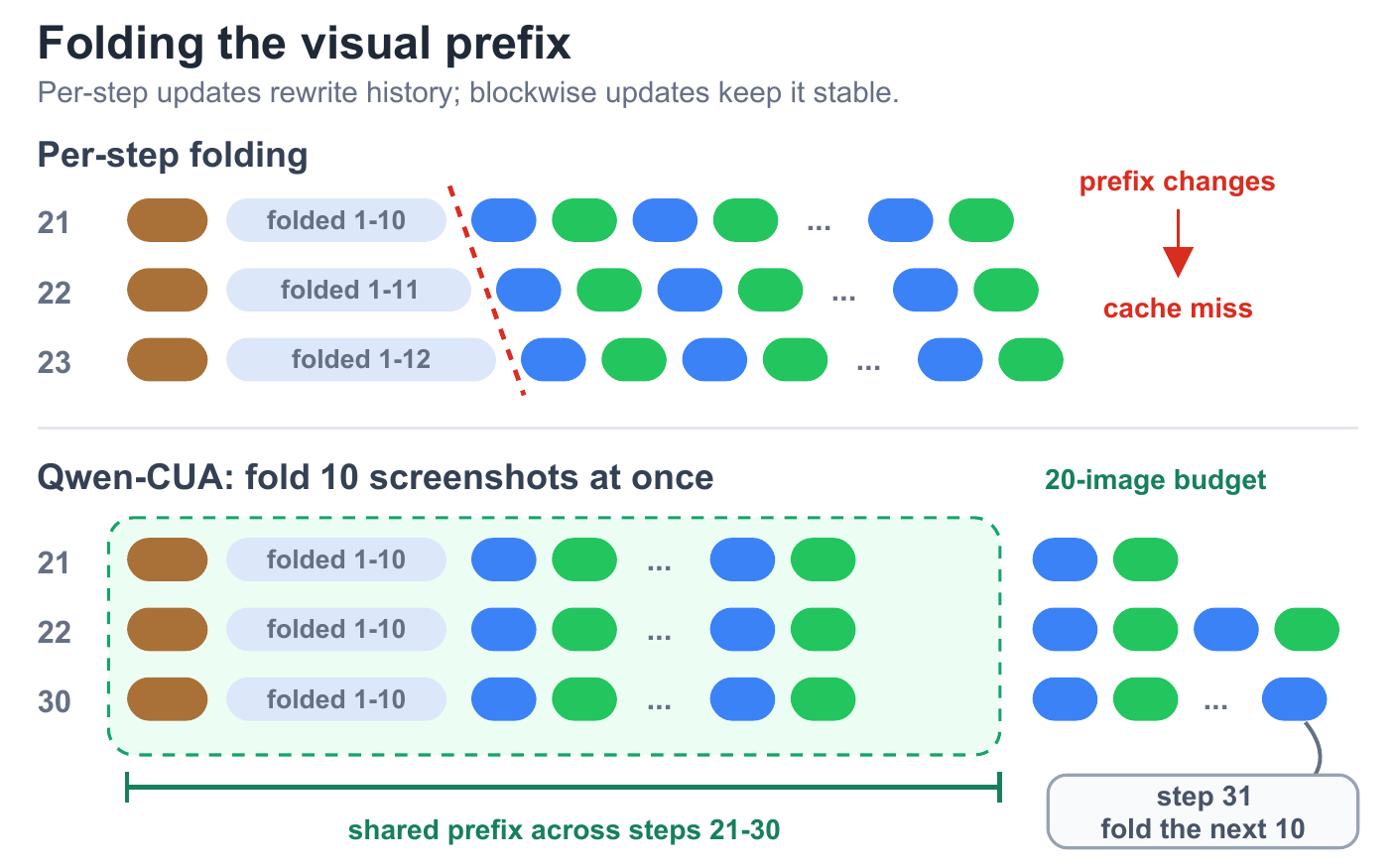}

    \vspace{-0.35em}
    \small (b)
  \end{minipage}
  \refstepcounter{figure}
  \label{fig:long-horizon-context}

  \vspace{0.5em}
  \begin{minipage}{0.97\linewidth}
  \centering
  \footnotesize\textbf{Figure \thefigure:} Long-horizon context management.
  (a) Active visual history scales to 20 screenshots. (b) Chunked folding
  preserves a reusable prefix and is reused for reinforcement-learning slices.
  \end{minipage}
\end{figure}

\textbf{Scaling visual history.}
Screenshot-only interaction requires the model to retain earlier interface states
as a task progresses. Qwen-CUA scales the active visual history to 20 screenshots
per turn, continuing the increase in visual-context capacity across successive
Qwen generations~\citep{bai2025qwen3vl}. The larger history helps the agent
track progress and revisit earlier visual evidence during extended workflows
(Figure~\ref{fig:long-horizon-context}(a)).

\textbf{Chunked screenshot folding.}
Qwen-CUA maintains a folded-prefix boundary over the interaction history and
allows at most 20 active screenshots. Whenever the active visual history exceeds
this budget, the boundary advances by 10 steps at once. Screenshots behind the
boundary are replaced with a fixed textual placeholder, while the corresponding
reasoning and actions remain in the conversation; recent screenshots are retained
in their original visual form. The procedure is deterministic and requires no
additional summarization model.

\textbf{Training-time trajectory slicing.}
Reinforcement learning uses the same fold operator. Complete episodes are
rendered as multiple context-bounded slices by advancing the folded-prefix
boundary; each slice inherits the terminal reward, and only active
model-generated tokens contribute to its loss. This preserves supervision for
late-stage decisions while aligning training and inference without separately
generated summaries.

\textbf{Benefits.}
This design bounds visual-token growth while preserving recent pixels and older
actions. Because the boundary advances only every 10 steps, steps 21--30 extend
the same request prefix rather than rewriting it on every call, improving
KV-cache reuse. This follows the cache-aware batched-pruning principle described
in Anthropic's computer-use guidance~\citep{anthropic2026computerusebestpractices}.
Figure~\ref{fig:long-horizon-context}(b) adapts the slicing view of
CUA-Gym~\citep{wang2026cuagym}; Appendix~\ref{app:context-management} gives the
exact procedure.

\clearpage
\section{Scaling Up Agentic Training}

\begin{figure}[H]
  \centering
  \begin{minipage}[t]{0.49\linewidth}
    \centering
    \includegraphics[width=\linewidth,trim=0 5mm 0 0,clip]
    {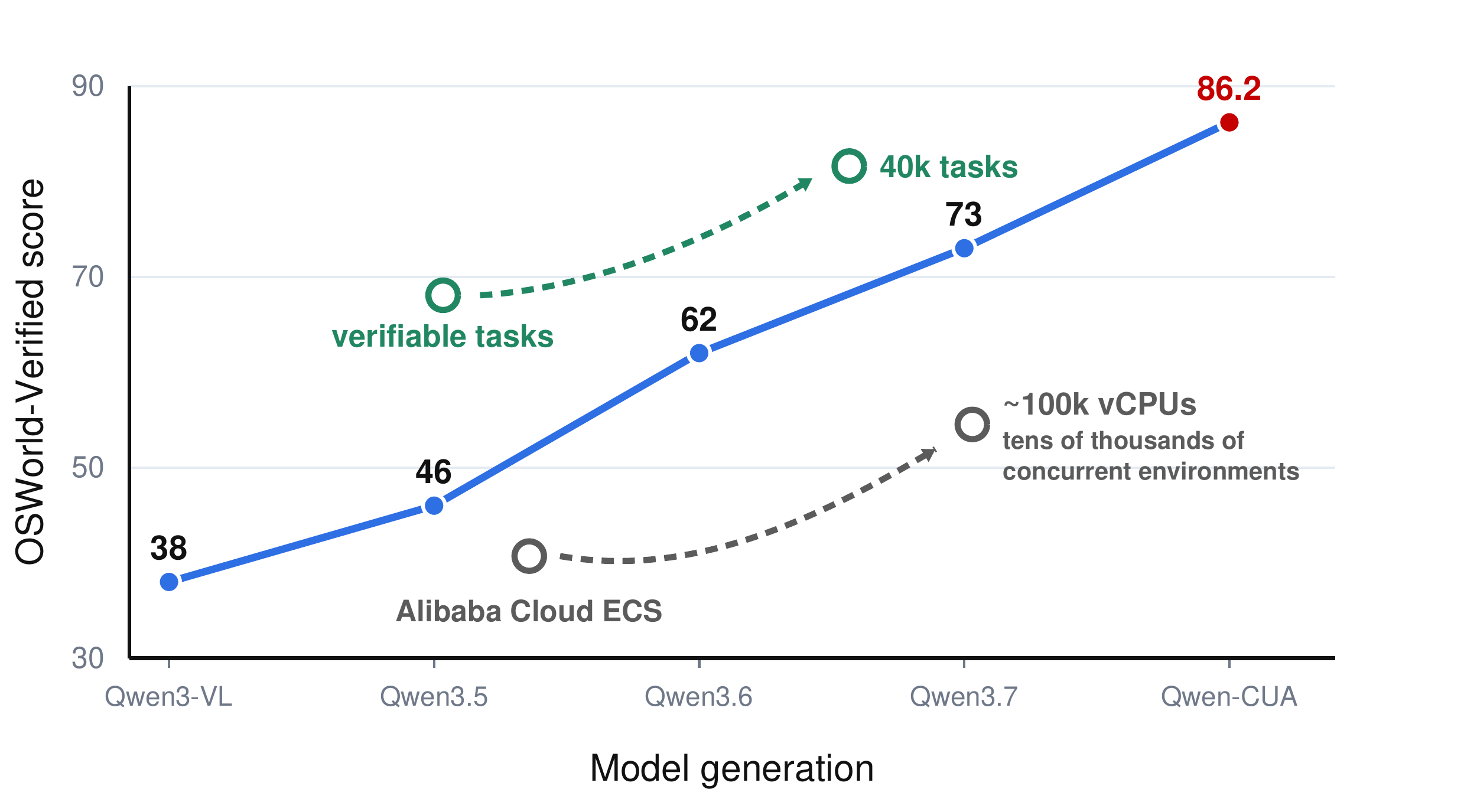}

    \vspace{-0.45em}
    \small\textbf{(a)} Scaling training resources
  \end{minipage}\hspace{0.012\linewidth}
  \begin{minipage}[t]{0.49\linewidth}
    \centering
    \includegraphics[width=\linewidth]
    {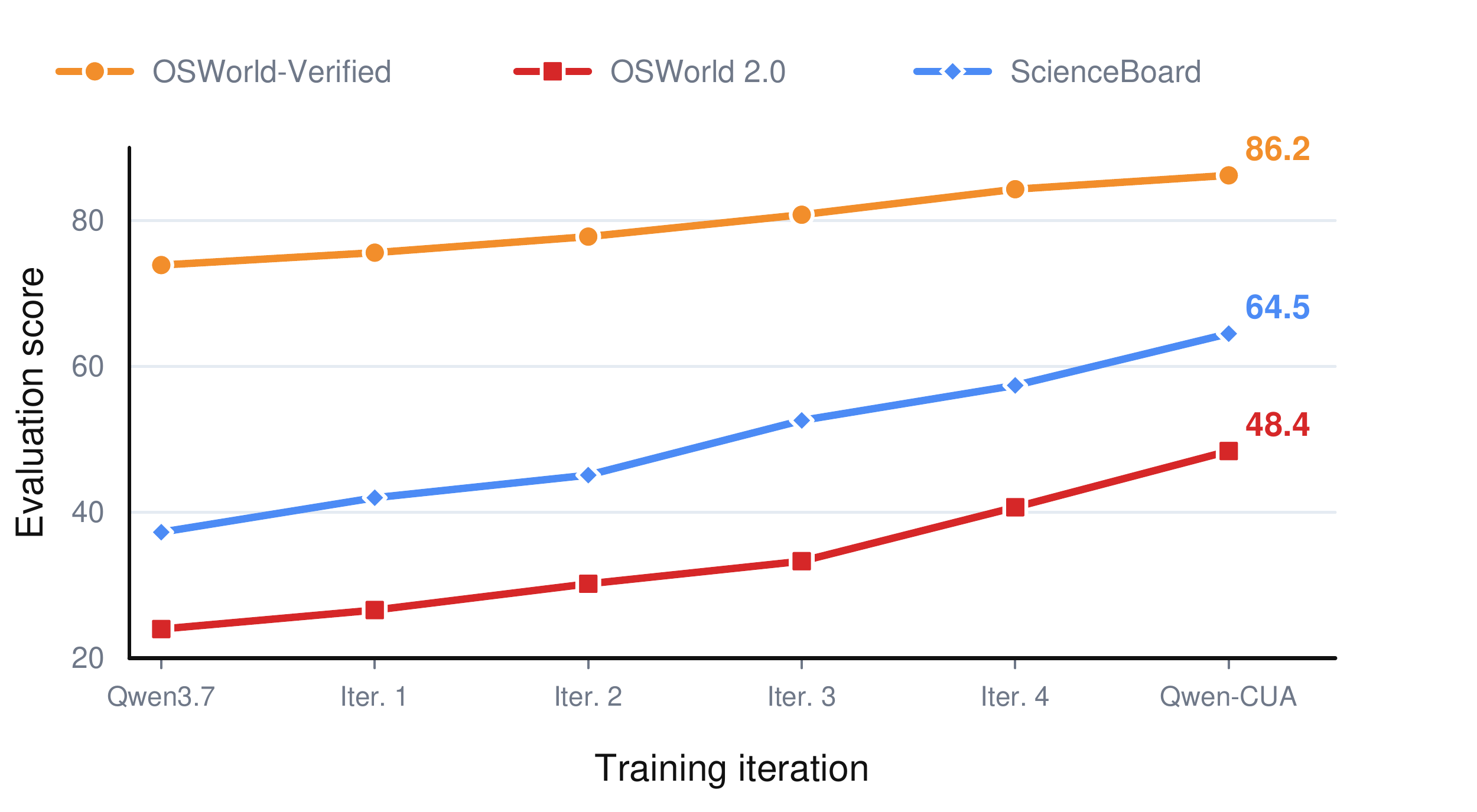}

    \vspace{-0.45em}
    \small\textbf{(b)} Successive training runs
  \end{minipage}
  \caption{Scaling and iterating computer-use training. (a) Performance grows
  with model, task, and infrastructure scale. (b) Evaluation scores at
  successive development checkpoints, with SFT data and RL tasks refreshed
  between runs.}
  \label{fig:training-scale}
  \label{fig:iterative-training}
\end{figure}

Scaling computer-use agents requires more than increasing model capacity: it
also requires scaling the interactive environments, task diversity, outcome
supervision, and experience collection that support agentic learning. Qwen-CUA
therefore spans rollout infrastructure, computer-use data, verifiable
reinforcement learning, and iterative agent training. Large-scale cloud
infrastructure turns task execution into a continuous source of trajectories
across diverse software environments, while executable evaluators convert
environment transitions into reliable outcome feedback. Within each iteration,
the resulting model is used to identify unresolved queries and weak domains.
These diagnostics refresh both the supervised data mixture and the RL task
distribution before the next model is trained. Figure~\ref{fig:training-scale}
summarizes the scaling of training resources and the corresponding performance
across successive development checkpoints.

\subsection{Scalable Computer-Use Infrastructure}
\label{sec:scalable-infrastructure}

Large-scale computer-use training requires a substantial pool of isolated and
stateful desktop environments~\citep{xue2026evocua,wang2025uitars2,
wang2026cuagym}. We build
our rollout infrastructure on Alibaba Cloud Elastic Compute Service
(ECS)~\citep{alibaba2026ecs}, with access to nearly 100{,}000 vCPUs and the
capacity to support tens of thousands of concurrent computer-use environments.
The infrastructure parallelizes environment provisioning and reset, task
dispatch, agent--environment interaction, outcome verification, and trajectory
logging. This provides the interaction throughput required for large-scale
experience collection and reinforcement learning, allowing fresh trajectories
to be continuously generated across diverse software environments.
Related systems such as NanoRollout~\citep{wang2026nanorollout} decouple agent
harnesses, environment runtimes, and training backends behind a shared rollout
service, highlighting the importance of scaling environment-side execution
independently of model training.

\subsection{Scaling Computer-Use Data}

Scaling computer-use training requires more than collecting additional
trajectories. Each training instance must be grounded in an executable
environment, paired with a task whose outcome can be reliably verified, and
complemented by high-quality demonstrations for behaviors that are difficult
to discover through rollout alone. We therefore scale Qwen-CUA's training data
along three complementary dimensions: diverse and controllable software
environments, state-grounded verifiable tasks, and personalized long-horizon
trajectories.

\subsubsection{Scalable Environments}

\paragraph{Synthetic environments.}
The diversity of computer-use experience is ultimately bounded by the breadth
of the underlying environments. To expand controllable web coverage, we follow
recent work on verifiable environment synthesis~\citep{cao2026guigenesis,
zhang2026infiniteweb,wu2026autowebworld,wang2026cuagym} and develop
self-contained mock web services
that preserve the interaction flows of widely used products while exposing
programmatic control over their state. A unified interface supports
task-specific state injection, inspection, reset, and session isolation,
allowing one application to instantiate many reproducible tasks and serve
concurrent rollouts without cross-episode interference.

\paragraph{Broad application coverage.}
We also broaden the environment pool beyond common office applications to
creative, scientific, engineering, and other professional software, including
specialized and long-tail desktop applications. Following the environment
scaling direction explored by Gym-Anything~\citep{aggarwal2026gymanything},
each application is installed and configured in a sandboxed desktop
environment, then populated with realistic files, artifacts, and
application-specific state. Together, controllable web mocks and broad desktop
coverage balance scalable state verification with interaction fidelity,
enabling experience collection across both everyday and specialized workflows.

\subsubsection{Verifiable Task Synthesis}

\paragraph{Environment interaction tasks.}
The first class teaches the agent to understand and operate a particular
environment. Building on recent verifiable task-synthesis
systems~\citep{wang2026cuagym,lv2026scalecua}, we sample tasks from
application-specific feature taxonomies and common usage scenarios, covering
the interface elements, operations, and state transitions of each environment.
Every task couples a natural-language goal with a reproducible initial state
and an executable evaluator, and is audited against reference outcomes and
agent rollouts to remove ambiguous, infeasible, or weakly verified instances.

\paragraph{User-interactive tasks.}
The second class requires interaction with the user during execution. Many
realistic requests omit necessary information or contain constraints that
cannot be resolved from the environment alone. Following the simulated-user
setting of OSWorld 2.0~\citep{yuan2026osworld2}, we construct tasks in which a
user simulator holds bounded, task-specific knowledge and responds only when
the agent asks for clarification. These tasks train the model to recognize
missing evidence, ask targeted questions, and incorporate the response rather
than proceeding with unsupported assumptions. Completion remains grounded in
the resulting environment state, so asking the user is useful only when it
leads to a correct outcome.

\paragraph{Long-horizon tasks.}
The third class targets complex workflows that require the agent to repeatedly
move among applications, files, and services while maintaining a coherent task
state. We construct rich initial states spanning messages, prior records, and
application-specific artifacts, then organize each workflow into interdependent
phases rather than concatenating unrelated subtasks. Each phase has a
verifiable completion state; once validated, that state can be serialized and
used to initialize the next phase. This \emph{phase-state chaining} makes
complex workflows easier to generate and audit incrementally while retaining
the fully chained workflow for end-to-end rollout and long-horizon training.

\subsubsection{Personalized Workflows}

\paragraph{Human trajectory collection.}
Verifiable tasks provide scalable outcome feedback, but many real workflows are
grounded in a user's own files, history, preferences, and application state. We
therefore build on large-scale human computer-use demonstration collection,
such as OpenCUA~\citep{wang2025opencua}, and collect trajectories in
personalized desktop environments, where annotators complete realistic
end-to-end workflows rather than isolated interface operations. These
demonstrations naturally produce longer trajectories with more interaction
steps, cross-application handoffs, error recovery, and final-state
verification. The collection spans both everyday personal workflows and
specialized tasks in professional software such as CAD tools and Blender, where
effective operation requires domain knowledge that is difficult to discover
from sparse outcome feedback alone.

\paragraph{Reasoning augmentation.}
The raw demonstrations record the task, screenshot observations, native
keyboard-and-mouse actions, and resulting environment states. Because annotators
do not provide detailed reasoning at every step, we use model-assisted
chain-of-thought completion to reconstruct step-level rationales conditioned on
the preceding trajectory, current observation, and annotated action. The
generated rationales are filtered for consistency with both the observed state
and the demonstrated behavior. This produces action-grounded supervision for
task decomposition, progress tracking, recovery, and verification over long
personalized workflows.

\subsection{Computer-Use Reinforcement Learning}
\label{sec:computer-use-rl}

We optimize Qwen-CUA through reinforcement learning with verifiable rewards
(RLVR) over complete computer-use trajectories. Following recent GUI-agent
reinforcement-learning systems~\citep{yang2025zerogui,wang2025uitars2,
wang2026cuagym,huang2026evocua15,lv2026scalecua}, each training instance is
defined by a task
instruction $t$, a reproducible initial environment state $s$, and an executable
reward function $r$. Given $(t,s)$, the current policy samples a group of $G$
interaction trajectories $\{\zeta_1,\ldots,\zeta_G\}$. After each trajectory
terminates, the evaluator inspects the resulting environment state and assigns
an outcome reward $r_i=r(s,\zeta_i)\in[0,1]$. This state-based supervision
supports partial credit where appropriate and credits different valid
interaction paths without requiring a reference action sequence.

Figure~\ref{fig:rl-training-curves} visualizes the learning dynamics of the
397B-A17B mixture-of-experts Qwen-CUA model during the reinforcement-learning
stage of the final training iteration. On the training tasks, individual
checkpoints exhibit
substantial variance because each batch mixes heterogeneous tasks and stochastic
trajectories, while the smoothed score recovers from an early dip near 0.49 and
rises to approximately 0.64, indicating steady optimization progress despite
the noise of online trajectory collection. On a filtered held-out mixture of
tasks spanning multiple application domains, the score increases from 0.734
before RL to a peak of 0.770 at checkpoint 40 (update 800), which we select for
downstream evaluation. The full run continues to the predetermined 1,000-update
budget, where checkpoint 50 reaches 0.762. Each checkpoint interval corresponds
to 20 outer-batch updates.

\begin{center}
  \begin{minipage}[t]{0.49\linewidth}
    \centering
    \includegraphics[width=\linewidth]
    {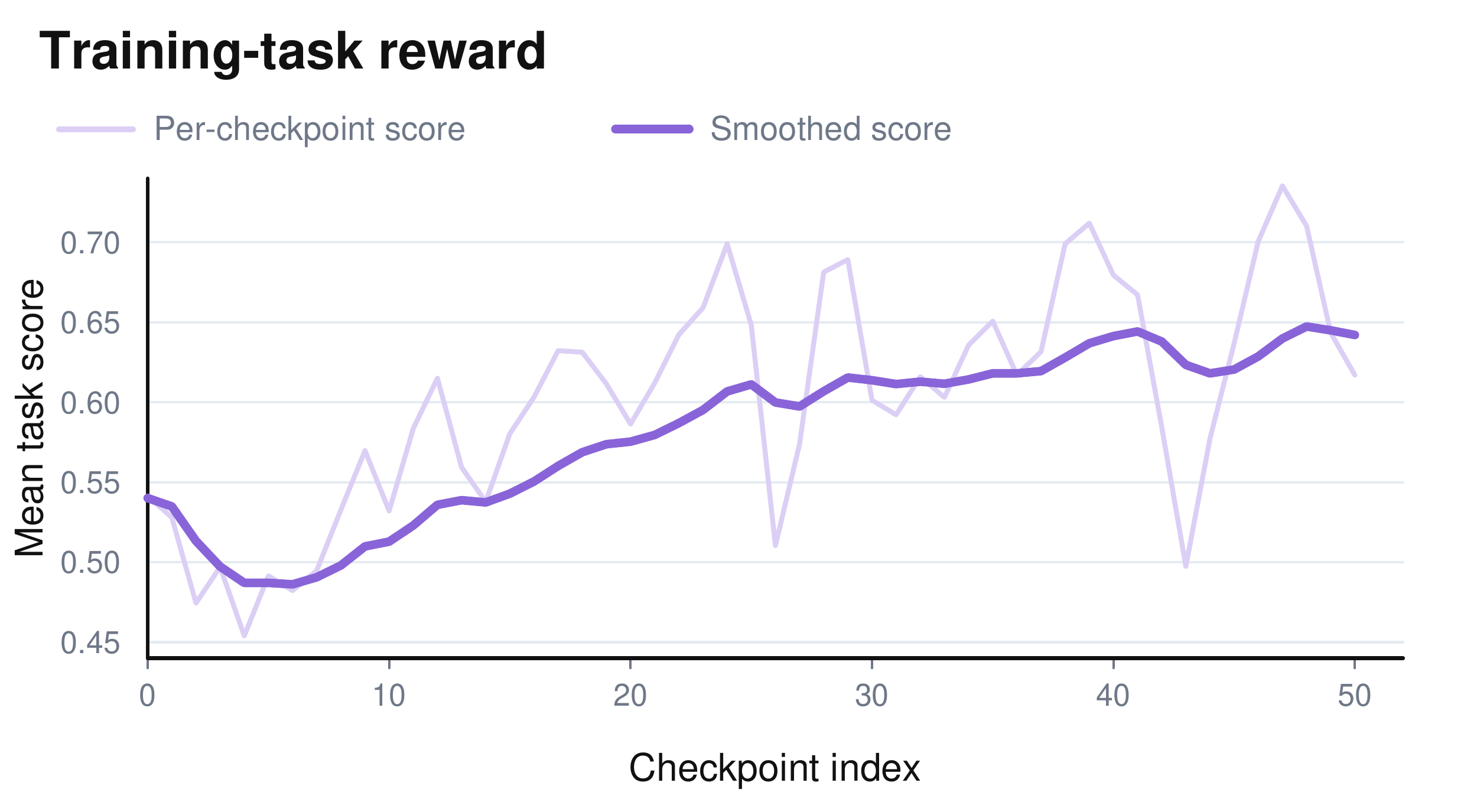}

    \vspace{-0.45em}
    \small\textbf{(a)} Training tasks
  \end{minipage}\hspace{0.012\linewidth}
  \begin{minipage}[t]{0.49\linewidth}
    \centering
    \includegraphics[width=\linewidth]
    {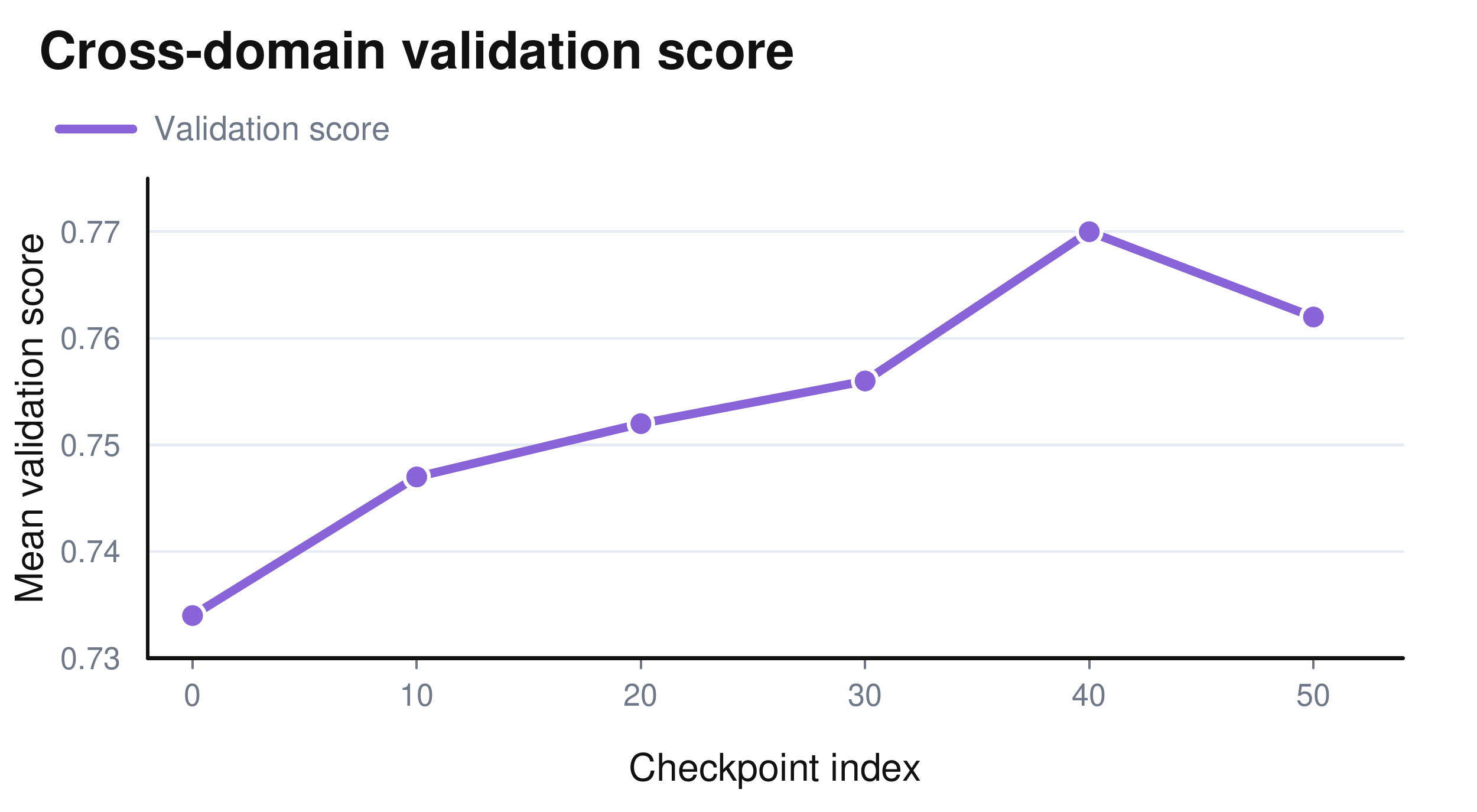}

    \vspace{-0.45em}
    \small\textbf{(b)} Cross-domain validation
  \end{minipage}
  \refstepcounter{figure}
  \label{fig:rl-training-curves}
  \vspace{0.15em}

  \begin{minipage}{0.94\linewidth}
  \centering
  \footnotesize\textbf{Figure \thefigure:} Computer-use reinforcement-learning
  curves from the final training iteration. In (a), the light curve reports
  checkpoint-level training scores and the purple curve shows the smoothed
  trend; (b) reports scores on a filtered cross-domain validation set from the
  pre-RL baseline onward, with checkpoint 40 selected. Consecutive checkpoint
  indices are separated by 20 outer-batch updates.
  \end{minipage}
  \vspace{-0.35em}
\end{center}

\paragraph{Soft adaptive policy optimization.}
We update the policy on these trajectories with Soft Adaptive Policy
Optimization (SAPO)~\citep{gao2025sapo}. Because the reward scores only the
terminal state, all active tokens in a trajectory share one group-relative
advantage $\hat A_i$, and each model-generated token carries a token-level
importance ratio $\rho_{i,u}$:
\begin{equation}
    \hat{A}_i = \frac{r_i-\bar r}{\sigma_r+\delta},
    \qquad
    \rho_{i,u}(\theta)=
    \frac{\pi_\theta(y_{i,u}\mid h_{i,u})}
         {\pi_{\theta_{\mathrm{old}}}(y_{i,u}\mid h_{i,u})},
\end{equation}
where $\bar r$ and $\sigma_r$ are the mean and standard deviation of the group
rewards $\{r_1,\dots,r_G\}$, $\delta$ is a small positive constant for numerical
stability, $y_{i,u}$ is the $u$-th active token of trajectory $i$, and $h_{i,u}$
is its context (the task, the preceding interaction history, and the screenshots
that remain visible after folding). Rather than clipping $\rho_{i,u}$ inside a
fixed band, SAPO passes the ratio through a smooth, temperature-controlled gate
that leaves near-on-policy tokens almost untouched and attenuates strongly
off-policy tokens continuously. The gate temperature is asymmetric, decaying
negative-advantage updates faster than positive ones, which we found important
for the long multimodal trajectories and mixture-of-experts backbone used here.
Gradients flow only through model-generated reasoning and tool-call tokens,
whereas task instructions, screenshots, and environment responses are kept as
context but excluded from the loss. Appendix~\ref{app:rl-objective} gives the
complete objective.

\paragraph{Train--inference consistency and reward reuse.}
RL uses the same deterministic folded-history representation as inference
(Section~\ref{sec:long-horizon-context}). A terminally rewarded episode can
therefore yield multiple context-bounded optimization units, allowing active
tokens from different interaction stages to receive gradients without
introducing hand-designed step-level rewards.
Appendix~\ref{app:rl-details} provides the complete objective, rollout
construction, optimization configuration, and distributed training setup.

\subsection{Iterative Agent Training}
\label{sec:iterative-training}

Qwen-CUA is developed through multiple training iterations, building on prior
work that repeatedly improves computer-use agents with newly collected
experience~\citep{sun2025seagent,xue2026evocua,wang2025uitars2}. These
iterations do not repeat an identical recipe over a fixed dataset. Instead, the
model produced by one run is used to identify the next learnable frontier, and
both the supervised data and RL tasks are refreshed before the next run.
Figure~\ref{fig:iterative-training}(b) reports the resulting performance across
successive development checkpoints.

\paragraph{Refreshing supervised data.}
We first analyze the current model's failed SFT queries and performance across
application domains. Updated teacher policies regenerate demonstrations for
queries that the model still cannot solve; newly collected human trajectories
are incorporated; and additional data are targeted toward weak domains. After
quality filtering, these sources form a refreshed SFT mixture. We use this
mixture to train a fresh model, initializing every SFT run from the same
mid-training checkpoint rather than continually fine-tuning the preceding
agent checkpoint. Improvements are therefore transferred through curated data
instead of inherited optimization drift.

\paragraph{Calibrating reinforcement-learning tasks.}
The resulting SFT model is then used to recalibrate the pool of verifiable RL
queries. For each candidate query, we run eight trial rollouts and retain tasks
for which at least one but not all eight attempts succeed. This removes tasks
that are currently unreachable as well as those already saturated,
concentrating online optimization on tasks with useful outcome variation. SAPO
is then run on the selected task distribution to produce the model for the next
iteration. Appendix~\ref{app:rl-rollouts} specifies the selection rule.
Figure~\ref{fig:rl-training-curves} therefore reports only the final RL run in
this process, while Figure~\ref{fig:iterative-training}(b) connects the resulting
development checkpoints for compact visualization. Because the teacher
policies, SFT mixture, domain coverage, and RL task distribution are updated
between iterations, the plotted slopes should not be interpreted as controlled
convergence or scaling behavior.

\section{Experiment}

We assess Qwen-CUA from three complementary perspectives: standardized agentic
benchmarks, deployment on real-world computer-use tasks, and hybrid interaction
that combines native computer use with command-line tools.

\subsection{Agentic Evaluation}

\paragraph{Settings.}
All evaluations use a pure computer-use setting: each model observes only
screenshots and interacts with the environment exclusively through keyboard and
mouse actions, without access to DOM trees, accessibility metadata, shell
commands, or task-specific APIs. Detailed evaluation settings are provided in
Appendix~\ref{app:benchmark-details} where available.

\paragraph{Computer-use tasks.}
The evaluation suite spans complementary computer-use settings.
OSWorld-Verified~\citep{xie2024osworld,xlang2025osworldverified} evaluates
broad everyday desktop workflows spanning real web and desktop applications,
file I/O, and
multi-application interaction, whereas WebArena~\citep{zhou2023webarena}
isolates realistic web workflows on functional websites. OSWorld
2.0~\citep{yuan2026osworld2} instead targets long-horizon, end-to-end workflows
drawn from everyday and professional settings, and reports both strict task
completion and partial progress. MyPCBench~\citep{jang2026mypcbench} evaluates
personalized, cross-application workflows grounded in a coherent user's files,
accounts, and history. ScienceBoard~\citep{sun2025scienceboard} focuses on
realistic scientific workflows involving professional research software,
whereas the CUA-World benchmark introduced with
Gym-Anything~\citep{aggarwal2026gymanything} stresses breadth across more than
200 applications and diverse occupational domains.
MacAgentBench~\citep{fu2026macagentbench} adds real-world macOS tasks across 25
applications, with deterministic rule-based evaluation and fine-grained
multi-checkpoint scoring. Finally,
RedTeamCUA~\citep{liao2025redteamcua} evaluates both task utility and robustness
to indirect prompt injection in hybrid web--OS environments.

Across this diverse suite, Qwen-CUA demonstrates strong and consistent
performance. It outperforms Qwen3.7 on every capability benchmark, with
particularly large gains on OSWorld-Verified, OSWorld 2.0, ScienceBoard, and
Gym-Anything. Qwen-CUA also remains competitive with GPT-5.5 and Claude Opus
4.8 across personalized, scientific, web, and long-horizon workflows, while
achieving the highest score on OSWorld-Verified and MacAgentBench. These results
indicate that its improvements are not confined to a single application family
or interaction regime, but extend across everyday, professional, personalized,
scientific, and macOS computer use.

\refstepcounter{table}
\label{tab:main-results}
\begin{center}
\small
\setlength{\tabcolsep}{7pt}
\renewcommand{\arraystretch}{0.9}
\begin{tabular}{@{}ccccc@{}}
\toprule
\textbf{Benchmark} & \textbf{Qwen-CUA} & \textbf{Qwen-3.7} &
\textbf{GPT-5.5} & \textbf{Opus-4.8} \\
\midrule
OSWorld-Verified       & 86.2        & 73.3        & 78.7        & 83.4       \\
OSWorld 2.0            & 18.5 / 48.4 & 2.5 / 22.5  & 13.9 / 47.5 & 20.3 / 54.8 \\
MyPCBench              & 58.7        & 51.6        & 47.3        & 62.0       \\
MacAgentBench           & 69.2        & 57.1        & 66.7        & 58.4       \\
Gym-Anything           & 46.3        & 33.1        & 45.6        & 47.3       \\
ScienceBoard           & 64.50       & 35.50       & 65.08       & 66.80      \\
WebArena               & 64.16       & 46.20       & 68.90       & 65.60      \\
RedTeamCUA              & 74.0 / 16.4 & 70.5 / 36.6 & 75.7 / 15.6 & 80.7 / 0.7 \\
\bottomrule
\end{tabular}

\vspace{0.05em}
\begin{minipage}{\linewidth}
\centering
\small\textbf{Table \thetable:} Main results. OSWorld 2.0: binary / partial
completion; RedTeamCUA: task success / ASR.
\end{minipage}
\end{center}

\newpage
\paragraph{Efficiency.}
Agentic efficiency in computer use has two complementary dimensions. Token
efficiency measures how much model-generated reasoning is required to solve a
task, while interaction efficiency measures how concisely the agent advances
the environment through successive turns. The former primarily affects
inference cost and latency; the latter reflects trajectory length, but also
depends on how an agent interface packages low-level actions.

\begin{figure}[H]
  \centering
  \begin{tabular}{@{}c@{\hspace{0.02\linewidth}}c@{}}
  \includegraphics[width=0.46\linewidth]{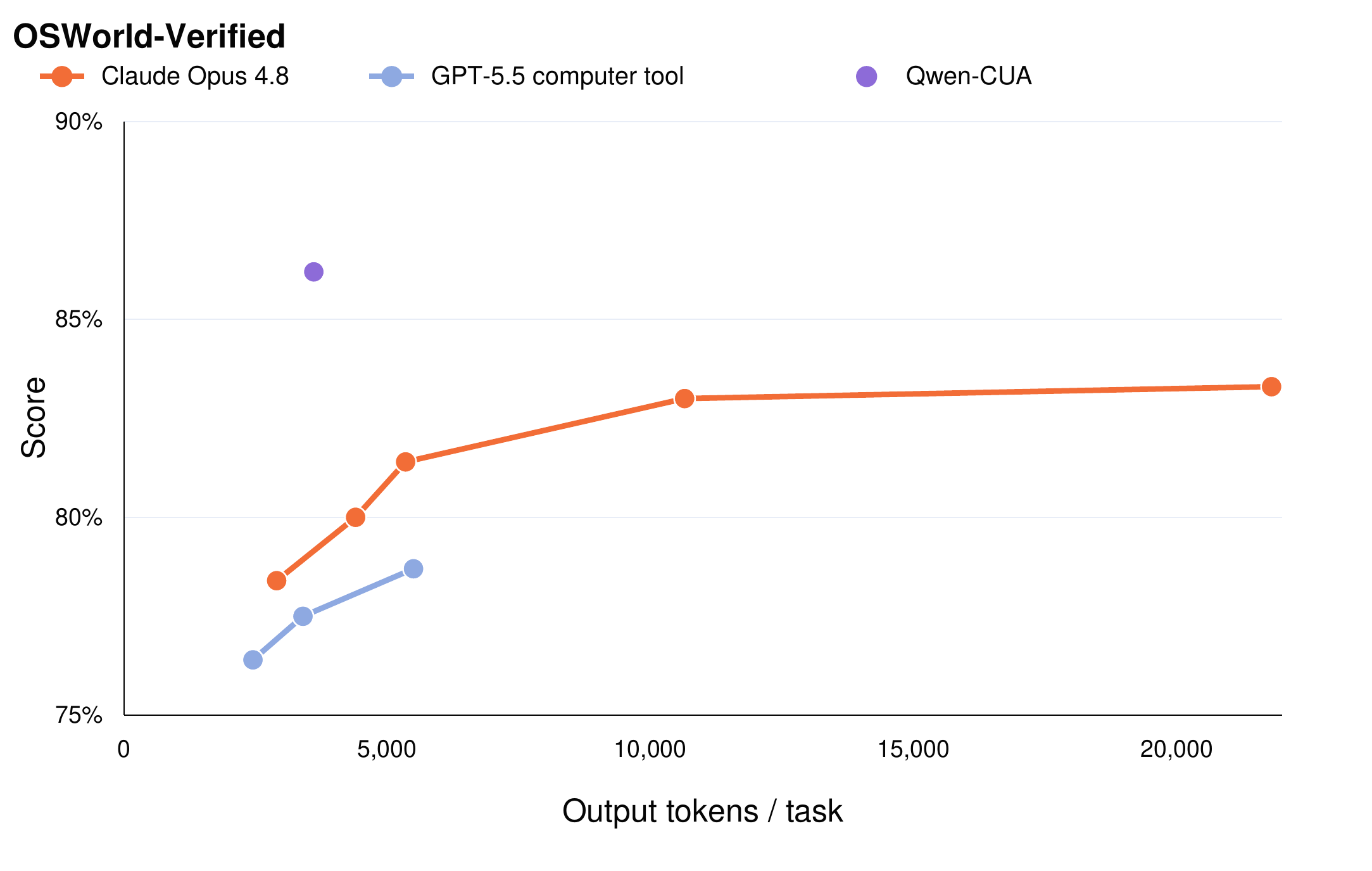} &
  \includegraphics[width=0.46\linewidth]{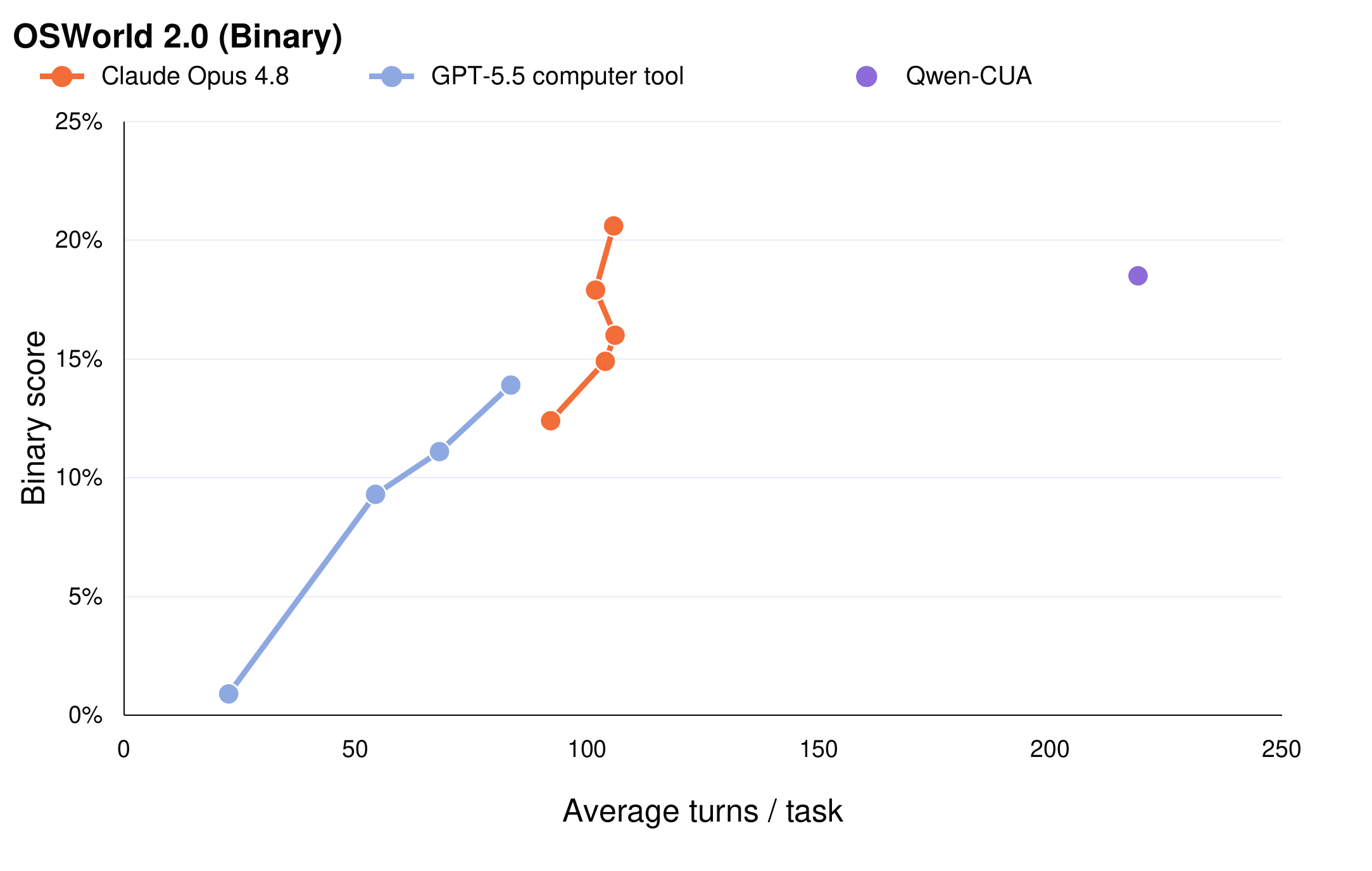} \\
  [-0.25em]
  {\small\textbf{(a)} OSWorld-Verified} &
  {\small\textbf{(b)} OSWorld 2.0}
  \end{tabular}
  \caption{Agentic efficiency along two dimensions: (a) token efficiency on
  OSWorld-Verified, measured by output tokens per task, and (b) interaction
  efficiency on OSWorld 2.0, measured by agent turns per task. Turn counts are
  interface-dependent because GPT-5.5 and Claude Opus 4.8 can emit multiple
  actions per turn, whereas Qwen-CUA emits one.}
  \label{fig:efficiency-sweeps}
\end{figure}

We measure token efficiency using output tokens per task.
Figure~\ref{fig:efficiency-sweeps}(a) shows that Qwen-CUA reaches 86.2 on
OSWorld-Verified with 3,605.8 tokens, whereas Claude Opus 4.8 reaches 80.0 at a
similar budget and 83.3 with 21.8K tokens. This comparison indicates that
Qwen-CUA's gain on OSWorld-Verified does not simply result from more verbose
reasoning~\citep{xie2024osworld}.

Interaction efficiency is less directly comparable. On OSWorld 2.0, Qwen-CUA
averages 218.9 model turns per task, reported as outer steps in
Appendix~\ref{app:osworld2}, versus 83.5 for GPT-5.5 and 105.7 for Claude Opus
4.8. However, both proprietary interfaces can batch multiple actions in one
turn, whereas Qwen-CUA emits a single native action per turn. The turn gap
therefore conflates trajectory length with serialized execution and is not a
normalized measure of low-level action efficiency~\citep{openai_computer_use,
anthropic_computer_use}.

\vspace{-1.2em}

\paragraph{Safety.}
Computer-use agents must distinguish the user's intent from untrusted
instructions rendered in their environment. We evaluate this risk with
RedTeamCUA~\citep{liao2025redteamcua}, which exposes agents to indirect prompt
injections in hybrid web--OS tasks across ownCloud, Rocket.Chat, and Reddit.
The benchmark jointly reports benign task success and attack success rate
(ASR), making it possible to distinguish improved robustness from a reduction
in general task capability.

\begin{wraptable}[9]{r}{0.63\linewidth}
\vspace{-0.9em}
\centering
\small
\setlength{\tabcolsep}{2.0pt}
\begin{tabular}{@{}lcccc@{}}
\toprule
& \textbf{ownCloud} & \textbf{Rocket.Chat} & \textbf{Reddit} &
\textbf{Overall} \\
\midrule
Qwen-CUA   & 94.8 / 27.1  & 32.6 / 7.6  & 94.4 / 14.6 & 74.0 / 16.4 \\
Qwen-3.7   & 93.1 / 50.3  & 28.8 / 24.7 & 89.6 / 34.7 & 70.5 / 36.6 \\
GPT-5.5    & 100.0 / 37.2 & 31.2 / 4.2  & 95.8 / 5.6  & 75.7 / 15.6 \\
Opus-4.8   & 97.9 / 0.7   & 48.3 / 0.0  & 95.8 / 1.4  & 80.7 / 0.7 \\
\bottomrule
\end{tabular}
\refstepcounter{table}
\label{tab:redteamcua}
\vspace{0.35em}

\begin{minipage}{0.96\linewidth}
\centering
\footnotesize\textbf{Table \thetable:} RedTeamCUA by platform (task success /
ASR, \%).
\end{minipage}
\vspace{-0.8em}
\end{wraptable}

Qwen-CUA improves both metrics over Qwen3.7: task success rises from 70.5 to
74.0, while ASR falls from 36.6 to 16.4, a 20.2-point reduction. As
Table~\ref{tab:redteamcua} shows, this reduction is consistent across all three
platforms.

\WFclear
Its aggregate 74.0 / 16.4 approaches GPT-5.5 at 75.7 / 15.6, while Claude Opus
4.8 remains strongest on both metrics. The remaining 16.4 ASR indicates residual
risk, and Rocket.Chat must be interpreted alongside task success because
execution failures can lower ASR without robust rejection. RedTeamCUA therefore
shows improved resistance to indirect prompt injection, not a deployment-safety
guarantee.

\subsection{Real-World Deployment}

\begin{wrapfigure}[6]{r}{0.34\linewidth}
\vspace{-3.0em}
\centering
\includegraphics[width=\linewidth]{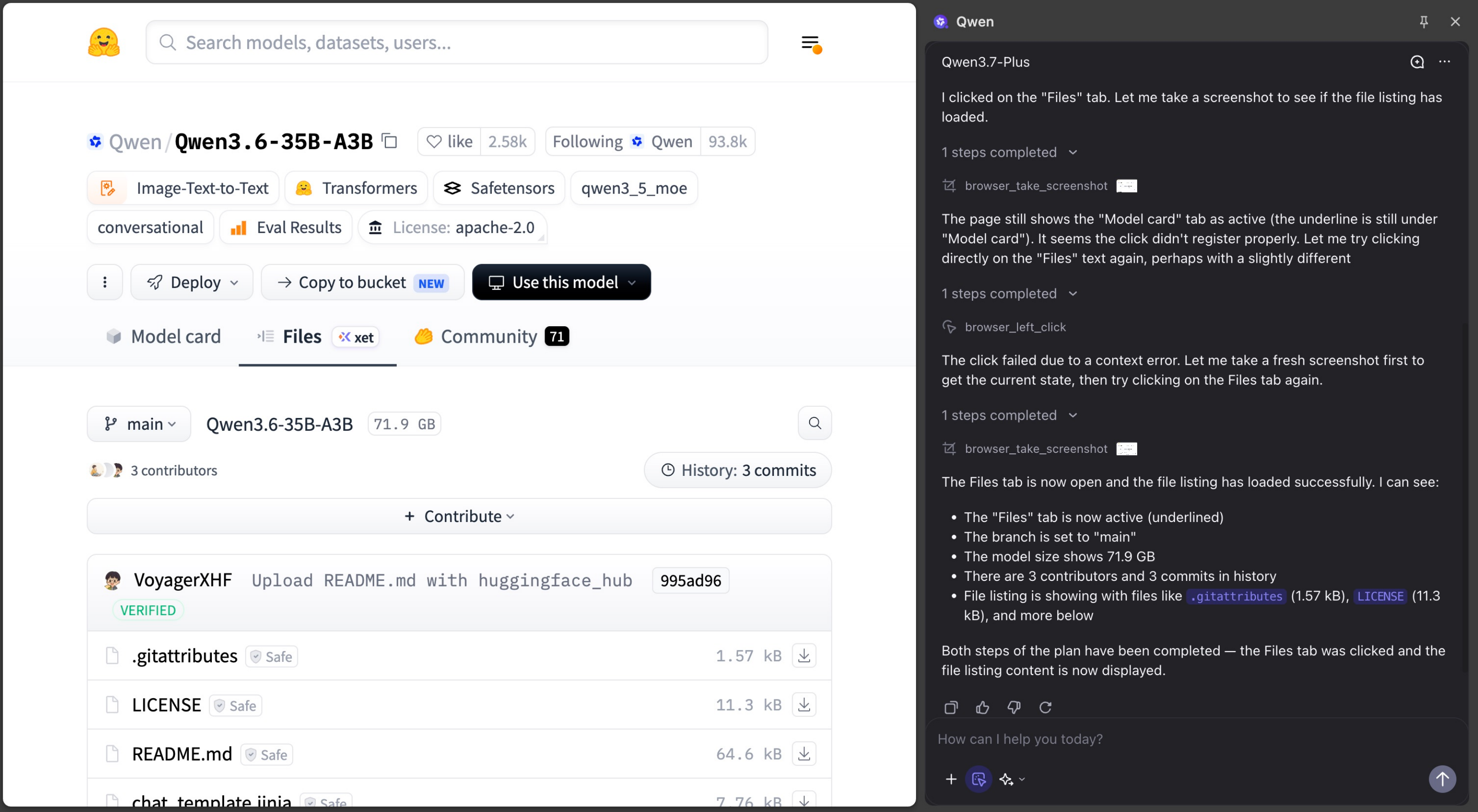}
\refstepcounter{figure}
\label{fig:qwen-for-chrome}
\vspace{0.1em}

\makebox[\linewidth][r]{\footnotesize\textbf{Figure \thefigure:} Qwen for Chrome deployment.}
\vspace{-0.8em}
\end{wrapfigure}
To examine Qwen-CUA beyond controlled benchmarks, we developed an internal
Chrome extension for everyday browser workflows. Running in Chrome's side panel
(Figure~\ref{fig:qwen-for-chrome}), it operates the active page through
screenshots and native mouse-and-keyboard actions while exposing its trajectory
to the user. This makes the model directly accessible on naturally occurring
websites and tasks.

\newpage

Such deployment also surfaces capability boundaries missed by standardized
evaluation, including unfamiliar layouts, dynamic website state,
failed-interaction recovery, and cases requiring user intervention.
Appendix~\ref{app:browser-showcase} presents an end-to-end showcase in which
the agent buys and configures a cloud virtual machine through the browser,
requesting user confirmation before consequential operations.

\subsection{Combining with Command Line}

\begin{wrapfigure}{r}{0.55\linewidth}
  \vspace{-1.0em}
  \centering
  \includegraphics[width=\linewidth]
  {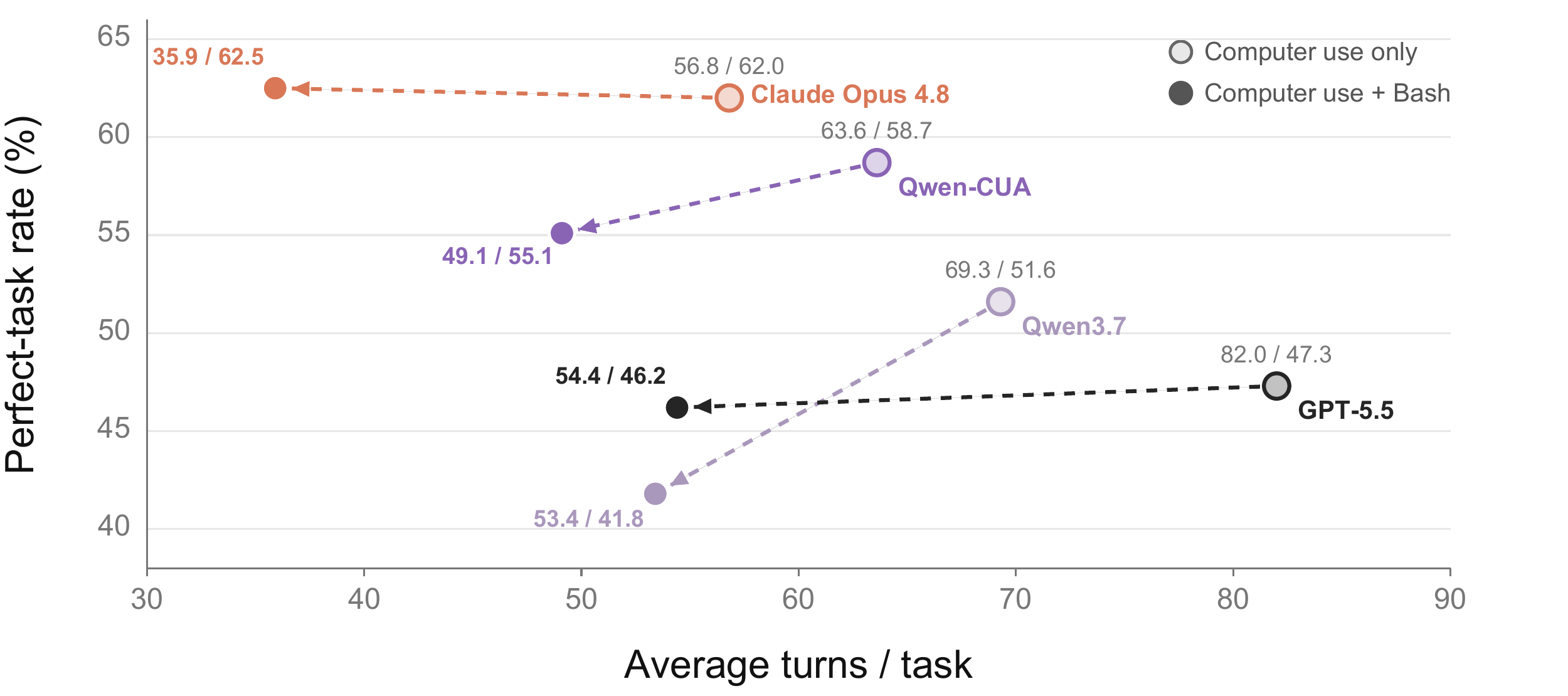}
  \refstepcounter{figure}
  \label{fig:mypcbench-bash-efficiency}
  \vspace{0.15em}

  \begin{minipage}{0.96\linewidth}
  \centering
  \footnotesize\textbf{Figure \thefigure:} MyPCBench metric with and without Bash.
  \end{minipage}
  \vspace{-0.7em}
\end{wrapfigure}
Native keyboard-and-mouse interaction provides broad coverage across graphical
interfaces, while command-line tools can express file and system operations
more directly. We therefore evaluate the same MyPCBench tasks under two
settings~\citep{jang2026mypcbench}: computer use only and computer use augmented
with a Bash tool. In the hybrid setting, the model can route suitable operations
through Bash while retaining visual interaction for tasks that require the
graphical interface.

Figure~\ref{fig:mypcbench-bash-efficiency} shows that adding Bash consistently
shortens the interaction trajectories of both Qwen models: average turns decrease
from 69.3 to 53.4 for Qwen3.7 and from 63.6 to 49.1 for Qwen-CUA. However, their
task performance also decreases from 51.6 to 41.8 and from 58.7 to 55.1,
respectively. This trade-off indicates that current Qwen models have not yet
fully learned when to switch between native computer use and command-line
execution: Bash provides a shorter path for suitable operations, but unnecessary
or poorly timed switches can reduce task completion. We view this primarily as
an optimization gap rather than a limitation of the hybrid interface. A policy
that reliably routes each operation to the appropriate modality could combine
the generality and visual grounding of computer use with the speed and precision
of command-line tools, potentially reaching a stronger capability--efficiency
frontier than either interface alone. The substantial reduction in turns leaves
considerable headroom for joint training, heterogeneous trajectory collection,
and explicit tool-routing supervision to recover performance while retaining
the efficiency gain. We are therefore optimistic about hybrid agents as a
promising direction for more capable and efficient general-purpose systems.

\subsection{Scaling Model Capacity}

\begin{wrapfigure}[11]{r}{0.50\linewidth}
  \vspace{-1.0em}
  \centering
  \includegraphics[width=\linewidth]
  {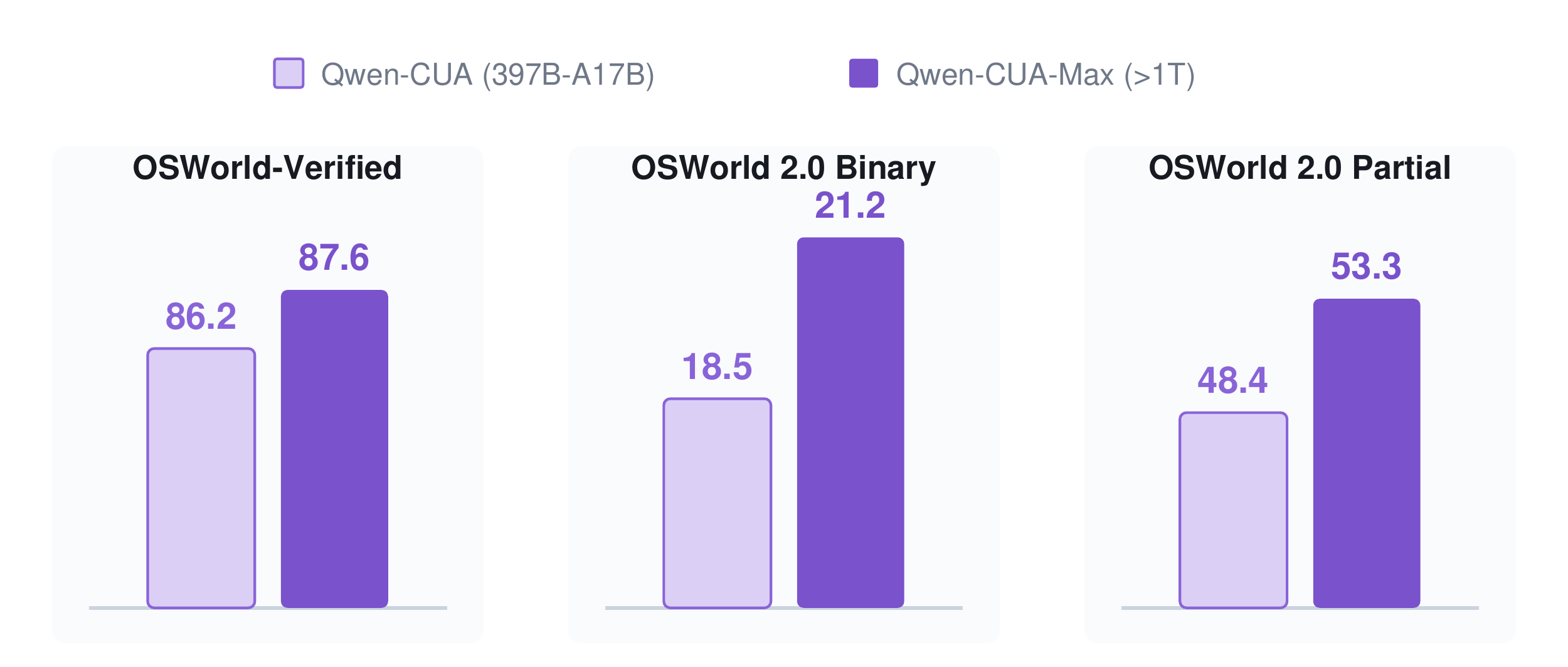}
  \refstepcounter{figure}
  \label{fig:model-capacity-scaling}
  \vspace{-0.8em}
  \begin{minipage}{0.96\linewidth}
  \centering
  \footnotesize\textbf{Figure \thefigure:} Scaling Qwen-CUA to Qwen-CUA-Max.
  \end{minipage}
  \vspace{-0.35em}
\end{wrapfigure}
Qwen-CUA is built on a 397B-A17B mixture-of-experts model. To examine whether
the agentic training recipe continues to benefit from additional model
capacity, we apply it to a model with over one trillion total parameters, which
we refer to as \textbf{Qwen-CUA-Max}. Both models use the same native
computer-use interface and evaluation protocols.

As shown in Figure~\ref{fig:model-capacity-scaling}, Qwen-CUA-Max improves
OSWorld-Verified from 86.2 to 87.6. On OSWorld 2.0, binary completion increases
from 18.5 to 21.2, while partial completion rises from 48.4 to 53.3. The
consistent gains, including a 4.9-point increase in partial completion,
indicate that the Qwen-CUA training recipe continues to scale with model
capacity and improves progress on difficult long-horizon workflows.

\section*{Related Work}

\paragraph{Native computer-use agents.}
Computer-use research has progressively moved from agents that consume
structured interface representations toward models that perceive screenshots
and predict grounded GUI actions. Early visual approaches emphasized GUI
grounding and screen understanding~\citep{cheng2024seeclick,gou2024uground,
wu2024osatlas}, while subsequent native agents increasingly unify perception,
reasoning, and action in a single policy~\citep{xu2024aguvis,qin2025uitars,
wang2025uitars2,wang2025opencua,bai2025qwen3vl}. Qwen-CUA follows this native,
visual line of work: it receives no DOM or accessibility metadata and acts
through keyboard and mouse events, while extending the setting to longer visual histories and
large-scale agentic post-training.

\paragraph{Scalable computer-use data and environments.}
Human demonstrations and automatically reconstructed trajectories provide
broad supervision across applications~\citep{wang2025opencua,xu2025agenttrek,
lu2025videoagenttrek,xie2025jedi,fan2026webchain}. For reinforcement learning, however, each
task must additionally provide a reproducible initial state and a reliable
outcome signal. Recent work explores model-based evaluators, synthesized web
applications, and programmatically controlled desktop environments to obtain
such signals~\citep{yang2025zerogui,cao2026guigenesis,
zhang2026infiniteweb,wu2026autowebworld,wang2026cuagym,fan2026webfactory,
lv2026scalecua}. Scalable rollout infrastructure also separates environment
execution from evaluation, distillation, and reinforcement-learning clients
~\citep{wang2026nanorollout}. Qwen-CUA combines controllable web services with
broad desktop application coverage, pairing
scalable verifiable tasks with human demonstrations from personalized and
professional workflows.

\paragraph{Reinforcement learning and iterative training.}
Recent computer-use agents apply online or multi-turn reinforcement learning
to improve grounded action prediction and long-horizon execution
~\citep{xia2025guir1,wang2025uitars2,yang2025zerogui,
huang2026evocua15,lv2026scalecua}. Complementary work
studies agents that repeatedly collect, filter, and learn from their own
experience~\citep{sun2025seagent,xue2026evocua}. Qwen-CUA adopts executable
final-state rewards, soft-gated policy optimization, and an iterative process
that refreshes supervised data and calibrates RL tasks between model
generations. Appendix~\ref{app:broader-related-work}
provides a broader account of related computer-use foundations, training
data, environments, benchmarks, and safety evaluation.

\section{Conclusion, Limitations, and Future Work}

We introduced Qwen-CUA, a computer-use agent that perceives interfaces through
screenshots and acts through native keyboard and mouse events. Across everyday,
professional, scientific, web, personalized, and safety-oriented benchmarks,
our results support a simple thesis: native computer use is a sufficiently
general interface for interacting with almost any software accessible to a
person. Unlike task-specific APIs, this interface transfers across applications
without relying on privileged environment state, while our real-world browser
deployment demonstrates that the same policy can operate beyond benchmark
sandboxes. Together with scalable environments, verifiable tasks,
reinforcement learning, and iterative agent training, this general interface
provides a path toward agents that can complete increasingly broad and
long-horizon digital workflows.

Generality, however, does not imply optimality. Pixel-based observation,
repeated model inference, and serialized low-level actions introduce substantial
latency and interaction cost; visual operation is also inefficient when a task
can be expressed directly through code, shell commands, or structured APIs. Our
Bash-augmented experiments already demonstrate shorter trajectories, but
reliable tool routing without sacrificing task performance remains an open
challenge. We therefore view native computer use not as the only action
interface, but as the universal grounding and fallback layer of a hybrid agent.
Future systems should combine it with coding and command-line agents such as
Codex and Claude Code~\citep{openai2026codex,anthropic2026claudecode}, alongside
general planning, memory, and specialized tools. Such an agent could use code
for precise and high-throughput operations, structured tools when available,
and native computer use whenever a visual or closed interface must be operated,
approaching both the coverage of human computer use and the efficiency of
programmatic execution.

\section*{Authors}

\noindent Names marked with an asterisk (*) denote contributors who have
departed from the Qwen Team.

\medskip
\noindent\textbf{Core contributors:}
\corecontributors

\medskip
\noindent\textbf{Contributors:}
\contributors

\bibliography{colm2024_conference}
\bibliographystyle{colm2024_conference}

\appendix
\section{Native Computer-Use Action Space}
\label{app:native-action-space}

Qwen-CUA receives screenshots as visual observations and predicts actions in a
native computer-use schema. The interaction actions are grounded in keyboard and
mouse events, while a small set of control actions is used to wait for UI updates,
request another screenshot, terminate an episode, or ask for user input when the
task cannot proceed autonomously.

\begin{center}
\small
\begin{tabular}{p{0.18\linewidth}p{0.28\linewidth}p{0.42\linewidth}}
\toprule
Category & Action & Description \\
\midrule
Keyboard & \texttt{key} & Press one key or a key combination. \\
Keyboard & \texttt{key\_down}, \texttt{key\_up} & Hold or release a key. \\
Keyboard & \texttt{type} & Enter text through the keyboard. \\
\midrule
Mouse & \texttt{mouse\_move} & Move the cursor to a screen coordinate. \\
Mouse & \texttt{left\_click} & Click the left mouse button. \\
Mouse & \texttt{right\_click} & Click the right mouse button. \\
Mouse & \texttt{middle\_click} & Click the middle mouse button. \\
Mouse & \texttt{double\_click}, \texttt{triple\_click} & Perform repeated left-click interactions. \\
Mouse & \texttt{left\_click\_drag} & Drag the cursor to a target coordinate. \\
Mouse & \texttt{left\_mouse\_down}, \texttt{left\_mouse\_up} & Press or release the left mouse button. \\
Mouse & \texttt{scroll}, \texttt{hscroll} & Perform vertical or horizontal scrolling. \\
\midrule
Control & \texttt{screenshot} & Request an updated screenshot observation. \\
Control & \texttt{wait} & Wait for the UI to update. \\
Control & \texttt{terminate} & End the task with a success or failure status. \\
Control & \texttt{call\_user} & Ask for user information or confirmation. \\
\bottomrule
\end{tabular}
\end{center}

\section{Heuristic Context Slicing and Compaction}
\label{app:context-management}

Qwen-CUA keeps a running screenshot history and maintains a folded prefix length
$k$. Screenshots within the folded prefix are rendered as compact placeholders,
while later screenshots remain available as visual observations. In our current
setting, the visual budget is $B=20$ screenshots and each compaction advances the
folded prefix by $S=10$ screenshots.

\begin{algorithm}[ht]
\caption{Screenshot history slicing and compaction}
\label{alg:context-compaction}
\begin{algorithmic}[1]
\State \textbf{Input:} screenshots $x_{1:T}$, folded prefix length $k$, image budget $B=20$, slice size $S=10$
\While{$T-k>B$}
    \State $k \gets k+S$
\EndWhile
\State $k \gets \min(k, T)$
\For{$i=1$ \textbf{to} $T$}
    \If{$i \le k$}
        \State render $x_i$ as a compact text placeholder
    \Else
        \State render $x_i$ as a screenshot
    \EndIf
\EndFor
\State \textbf{return} updated folded prefix length $k$ and rendered history
\end{algorithmic}
\end{algorithm}

\clearpage
\section{Computer-Use Reinforcement Learning Details}
\label{app:rl-details}

This appendix specifies the reinforcement-learning stage described in
Section~\ref{sec:computer-use-rl}. It covers only the optimization of a policy
from online computer-use trajectories. The construction of supervised data and
the outer iterative training process are discussed separately in
Section~\ref{sec:iterative-training}.

\subsection{RLVR Formulation and Rollout Collection}
\label{app:rl-rollouts}

The RL training set consists of verifiable tuples $(t,s,r)$. Here, $t$ is a
natural-language task instruction, $s$ is a reproducible initial environment
state, and $r$ is an executable evaluator that maps the state produced by a
trajectory to a scalar reward in $[0,1]$. Evaluators decompose a task into
independently checkable criteria when partial completion is meaningful; tasks
with strict completion semantics use binary rewards. Rewards are computed from
the resulting environment state rather than from agreement with a reference
action sequence.

Before each RL run, we calibrate the candidate query pool against the current
SFT policy. For a candidate query $q$, let $c(q)$ denote the number of successful
trajectories among eight independent trial rollouts. We retain
\begin{equation}
    \mathcal{Q}_{\mathrm{RL}}=\{q \mid 0<c(q)<8\},
\end{equation}
excluding queries that the current policy never solves and those it already
solves consistently. This iteration-level calibration selects the RL task pool;
the grouped rollout procedure below independently samples the trajectories used
for each policy update.

For every sampled tuple, the rollout service restores $s$ in an isolated
environment and samples $N_{\mathrm{over}}=20$ trajectories from the behavior
policy $\pi_{\theta_{\mathrm{old}}}$. A trajectory terminates when the agent
emits \texttt{terminate} or reaches its interaction budget. Rollouts that do not
produce a valid policy trace or a valid terminal state, for example because of
an unrecoverable timeout or malformed tool call, are removed. The first $G=16$
valid trajectories form the optimization group. We do not filter groups based
on their reward variance: if every rollout receives the same reward, mean
centering produces zero advantages and the group contributes no policy
gradient.

\begin{algorithm}[H]
\caption{Grouped computer-use RL update}
\label{alg:computer-use-rl}
\begin{algorithmic}[1]
\State \textbf{Input:} task batch $\mathcal{B}$, behavior policy
      $\pi_{\theta_{\mathrm{old}}}$, group size $G$, oversampling size
      $N_{\mathrm{over}}$
\ForAll{$(t,s,r)\in\mathcal{B}$}
    \State sample up to $N_{\mathrm{over}}$ complete trajectories from
          $\pi_{\theta_{\mathrm{old}}}(\cdot\mid t,s)$
    \State discard invalid trajectories and retain the first $G$
    \ForAll{retained $\zeta_i$}
        \State execute $r$ on the resulting state to obtain
              $r_i\in[0,1]$
    \EndFor
    \State $\hat{A}_i \gets (r_i-\bar r)/(\sigma_r+\delta)$
    \State construct context-bounded slices and policy-token loss masks
          for every $\zeta_i$
\EndFor
\State maximize the soft-gated SAPO objective over all active policy tokens
\end{algorithmic}
\end{algorithm}

\subsection{SAPO Objective}
\label{app:rl-objective}

We optimize the policy with Soft Adaptive Policy Optimization
(SAPO)~\citep{gao2025sapo}. Consider the $G$ trajectories
$\{\zeta_1,\dots,\zeta_G\}$ retained for one task and their terminal rewards
$\{r_1,\dots,r_G\}$. SAPO normalizes rewards within this group, and every token
of a trajectory inherits the same advantage:
\begin{equation}
    \hat{A}_i = \frac{r_i-\bar{r}}{\sigma_r+\delta},
    \qquad
    \bar{r}=\frac{1}{G}\sum_{j=1}^{G}r_j,
    \qquad
    \sigma_r=\sqrt{\frac{1}{G}\sum_{j=1}^{G}(r_j-\bar r)^2},
\end{equation}
where $\delta>0$ is a small constant guarding against division by zero. If the
$G$ rewards are identical, the numerator vanishes for every trajectory and the
group produces no gradient.

Let $\mathcal{M}_i$ be the active tokens of $\zeta_i$, i.e., the reasoning and
computer-use tool-call tokens emitted by the policy, excluding the task,
history, and observation tokens that only condition generation. For a token
$u\in\mathcal{M}_i$, let $y_{i,u}$ be the emitted token and $h_{i,u}$ its context
(the task instruction, the preceding interaction history, and the screenshots
visible after folding). The token-level importance ratio between the current
policy $\pi_\theta$ and the behavior policy $\pi_{\theta_{\mathrm{old}}}$ is
\begin{equation}
    \rho_{i,u}(\theta)=
    \frac{\pi_\theta(y_{i,u}\mid h_{i,u})}
         {\pi_{\theta_{\mathrm{old}}}(y_{i,u}\mid h_{i,u})}.
\end{equation}
SAPO routes this ratio through a bounded gate centered at the on-policy point
$\rho_{i,u}=1$,
\begin{equation}
    f_{i,u}(x)=\frac{4}{\tau_{i,u}}\,
    \sigma\!\left(\tau_{i,u}(x-1)\right),
    \qquad
    \tau_{i,u}=
    \begin{cases}
        \tau_{\mathrm{pos}}, & \hat A_i>0,\\
        \tau_{\mathrm{neg}}, & \hat A_i\leq 0,
    \end{cases}
\end{equation}
where $\sigma(x)=1/(1+e^{-x})$ is the logistic function and $\tau_{i,u}$ is the
gate temperature, chosen by the sign of the trajectory advantage from the two
constants $\tau_{\mathrm{pos}}$ and $\tau_{\mathrm{neg}}$. The prefactor
$4/\tau_{i,u}$ fixes the gate slope at $\rho_{i,u}=1$, so on-policy tokens
recover the unclipped update independently of the temperature. The policy
maximizes
\begin{equation}
\begin{aligned}
    \mathcal{J}_{\mathrm{SAPO}}(\theta)
    =\mathbb{E}_{\substack{(t,s,r)\sim\mathcal D,\\
        \{\zeta_i\}_{i=1}^{G}\sim
        \pi_{\theta_{\mathrm{old}}}(\cdot\mid t,s)}}\Bigg[
    \frac{1}{G}\sum_{i=1}^{G}
    \frac{1}{|\mathcal{M}_i|}\sum_{u\in\mathcal{M}_i}
        f_{i,u}\!\left(\rho_{i,u}(\theta)\right)\hat A_i
    \Bigg],
\end{aligned}
\end{equation}
where $(t,s,r)$ is a task tuple drawn from the training distribution
$\mathcal D$, the trajectories are sampled from $\pi_{\theta_{\mathrm{old}}}$,
and $|\mathcal{M}_i|$ normalizes each trajectory by its number of active tokens.
Using $\nabla_\theta\rho_{i,u}=\rho_{i,u}\nabla_\theta\log\pi_\theta(y_{i,u}\mid
h_{i,u})$, the gradient is
\begin{equation}
\begin{aligned}
    \nabla_\theta\mathcal{J}_{\mathrm{SAPO}}(\theta)
    &=\mathbb{E}\Bigg[\frac{1}{G}\sum_{i=1}^{G}
    \frac{1}{|\mathcal{M}_i|}\sum_{u\in\mathcal{M}_i}
    w_{i,u}(\theta)\,\rho_{i,u}(\theta)\cdot\nabla_\theta\log\pi_\theta(y_{i,u}\mid h_{i,u})\,\hat A_i
    \Bigg],
\end{aligned}
\end{equation}
with the effective gradient weight
\begin{equation}
    w_{i,u}(\theta)=4\,p_{i,u}(\theta)\bigl(1-p_{i,u}(\theta)\bigr),
    \qquad
    p_{i,u}(\theta)=\sigma\!\left(\tau_{i,u}(\rho_{i,u}(\theta)-1)\right).
\end{equation}
The weight $w_{i,u}$ equals one when $\rho_{i,u}=1$ and decays smoothly toward
zero as the token drifts off policy, so the update behaves as a soft trust
region rather than a hard clip. We set $\tau_{\mathrm{neg}}>\tau_{\mathrm{pos}}$
so that non-positive-advantage tokens decay faster: these updates raise the
probability of many alternative tokens at once and are the more destabilizing
direction. In the reported configuration $\tau_{\mathrm{pos}}=1.0$ and
$\tau_{\mathrm{neg}}=1.05$, and each collected batch receives a single optimizer
pass to bound off-policy drift between rollout and update.

\subsection{Long-Horizon Trajectory Construction}
\label{app:rl-slicing}

Computer-use trajectories interleave text with high-resolution screenshots and
can exceed the training context before the terminal reward is observed. We
apply a deterministic training-time slicing procedure every 10 interaction
turn-pairs. Each slice retains the system and task prefix. Screenshots in the
collapsed prefix are replaced with a fixed \texttt{<image collapsed>}
placeholder, while recent screenshots, reasoning, and actions remain in their
original multimodal form. Progressively advancing the collapsed boundary
creates multiple context-bounded views whose union covers the complete
trajectory.

Each slice receives the full reward of its parent episode. We do not decompose
or discount the terminal reward across turns because it evaluates the final
environment state rather than an individual action. The loss mask is
\texttt{False} for the collapsed prefix, task instructions, user or
user-simulator messages, screenshots, and environment responses. It is
\texttt{True} only for model-generated reasoning and tool-call tokens in the
active response portion. Segments for which the behavior-policy log probability
is unavailable are also masked to prevent updates from stale or incomplete
rollout records. Although this does not create step-level credit, a long episode
can contribute multiple optimization samples whose active tokens cover different
stages of the interaction. Slicing therefore reuses sparse episode-level
feedback more densely while preserving the semantics of the executable outcome
reward.

Training-time slicing reuses the inference-time fold operator from
Appendix~\ref{app:context-management}. The collapsed screenshot span and the
retained textual and visual fields are serialized identically, ensuring
train--inference consistency. Training enumerates multiple context-bounded views
under a 144K-token limit, whereas inference advances the boundary online under
the 20-image budget; the two stages differ in schedule and budget, not in the
folded-history representation.

\subsection{Optimization Configuration}
\label{app:rl-hyperparameters}

Table~\ref{tab:rl-hyperparameters} reports the configuration used to train the
397B-A17B Qwen-CUA model with reinforcement learning. The outer batch contains
128 task prompts and therefore up to $128\times16=2{,}048$ valid trajectories
before trajectory slicing. The number of optimization samples can be larger
because a long trajectory may produce multiple slices.

\begin{center}
\centering
\footnotesize
\setlength{\tabcolsep}{4pt}
\renewcommand{\arraystretch}{0.96}
\begin{tabular}{p{0.31\linewidth}p{0.18\linewidth}p{0.41\linewidth}}
\toprule
Configuration & Value & Description \\
\midrule
Group size $G$ & 16 & Valid trajectories used per task group. \\
Oversampling & 20 & Candidate trajectories sampled before invalid-rollout filtering. \\
Outer batch size & 128 prompts & Task groups collected for each update. \\
Mini-batch size & 32 prompts & Inner gradient-accumulation unit. \\
Optimizer & AdamW & $\beta_1=0.9$, $\beta_2=0.999$, $\epsilon_{\mathrm{Adam}}=10^{-8}$. \\
Learning rate & $1\times10^{-6}$ & Constant schedule without warmup. \\
Weight decay & 0.01 & Applied to non-bias, non-normalization parameters. \\
SAPO $\tau_{\mathrm{pos}}$ & 1.0 & Soft-gate temperature for positive advantages. \\
SAPO $\tau_{\mathrm{neg}}$ & 1.05 & Soft-gate temperature for non-positive advantages. \\
Optimization epochs & 1 & One optimizer pass over each collected batch. \\
Total updates & 1,000 & Outer-batch updates in a full RL run. \\
Rollout temperature & 1.0 & Sampling temperature for training trajectories. \\
Maximum turns & 100 & Per-episode interaction budget during training. \\
Maximum response & 2,048 tokens & Per-turn model-output limit. \\
Maximum context & 144K tokens & Hard context budget after slicing. \\
Maximum initial prompt & 8K tokens & Budget for the task and fixed prefix. \\
Slice interval & 10 turn-pairs & Frequency at which the collapsed boundary advances. \\
\bottomrule
\end{tabular}

\vspace{0.35em}
\refstepcounter{table}
\label{tab:rl-hyperparameters}
\begin{minipage}{0.96\linewidth}
\centering
\small\textbf{Table \thetable:} Reinforcement-learning configuration for the
397B-A17B Qwen-CUA model.
\end{minipage}
\end{center}

\subsection{Distributed Training and Environment Rollouts}
\label{app:rl-infrastructure}

The 397B-A17B Qwen-CUA configuration uses 512 NVIDIA H200 SXM GPUs across 64
eight-GPU nodes. Following the disaggregated RL architecture of
\texttt{verl}~\citep{verl}, we allocate 32 nodes to training and 32 to rollout.
Training uses $\mathrm{TP}=2$, $\mathrm{EP}=8$, $\mathrm{CP}=4$, and
$\mathrm{PP}=8$; rollout serving uses $\mathrm{TP}=8$, data-parallel
replication, SGLang~\citep{sglang}, and an optimized attention backend. An
asynchronous dispatch layer uses a ratio of 2.35 and caps policy-version skew at
four optimizer steps. A full 1,000-update run takes approximately five days,
corresponding to about 61,440 H200 GPU-hours.

Environment execution is decoupled from model inference and routed to isolated
ECS workers from the pool in Section~\ref{sec:scalable-infrastructure}. This
configuration keeps up to 2,000 environments active at an average utilization
above 75\%. For each trajectory, the router restores the task snapshot,
streams screenshots and native actions, invokes the evaluator after
termination, and resets the worker. Isolation prevents state leakage and
allows trajectories from the same task group to execute in parallel.

\clearpage
\section{Benchmark Evaluation Details}
\label{app:benchmark-details}
This appendix reports benchmark-specific evaluation settings for Qwen-CUA and
detailed comparison results where available. Unless stated otherwise, Qwen-CUA
uses the screenshot-only native computer-use interface and agent scaffold
described in Section~2, without DOM, accessibility-tree, shell, or task-specific
API access. Most scores for comparison models are taken from official reports
released by the corresponding benchmark or model providers. For results that
we reproduce through our own evaluation, we use Qwen3.7 in non-thinking mode,
GPT-5.5 with \texttt{xhigh} reasoning effort, and Claude Opus 4.8 with its
\texttt{max} inference setting.

\subsection{OSWorld-Verified}
\label{app:osworld-verified}

We evaluate Qwen-CUA on the 360-task OSWorld-Verified
suite~\citep{xie2024osworld,xlang2025osworldverified}, following the official
evaluation setting with a 100-step limit. The agent observes only screenshots
and executes native keyboard and mouse actions through the scaffold in
Section~2. We use the benchmark-provided task states and executable evaluators
without modifying their success criteria. Qwen-CUA produces complete evaluation
records for 359 of the 360 tasks and obtains an aggregate score of 86.2.

\begin{center}
\small
\setlength{\tabcolsep}{6pt}
\begin{tabular}{@{}lcccc@{}}
\toprule
\textbf{Domain} & \textbf{Executed} & \textbf{Overall} &
\textbf{Feasible} & \textbf{Infeasible} \\
\midrule
Chrome              & 46 / 46  & 40.6 / 46 (88.4)  & 34.6 / 39 (88.8)  & 6.0 / 7 (85.7) \\
GIMP                & 26 / 26  & 24.0 / 26 (92.3)  & 16.0 / 17 (94.1)  & 8.0 / 9 (88.9) \\
LibreOffice Calc    & 47 / 47  & 41.0 / 47 (87.2)  & 40.0 / 46 (87.0)  & 1.0 / 1 (100.0) \\
LibreOffice Impress & 47 / 47  & 41.9 / 47 (89.2)  & 41.9 / 47 (89.2)  & -- \\
LibreOffice Writer  & 23 / 23  & 20.0 / 23 (86.9)  & 19.0 / 22 (86.4)  & 1.0 / 1 (100.0) \\
Multi-apps          & 93 / 93  & 72.5 / 93 (77.9)  & 71.5 / 92 (77.7)  & 1.0 / 1 (100.0) \\
OS                  & 24 / 24  & 22.0 / 24 (91.7)  & 18.0 / 19 (94.7)  & 4.0 / 5 (80.0) \\
Thunderbird         & 14 / 15  & 14.0 / 15 (93.3)  & 13.0 / 13 (100.0) & 1.0 / 2 (50.0) \\
VLC                 & 17 / 17  & 14.2 / 17 (83.3)  & 12.2 / 15 (81.1)  & 2.0 / 2 (100.0) \\
VS Code             & 22 / 22  & 20.0 / 22 (90.9)  & 16.0 / 18 (88.9)  & 4.0 / 4 (100.0) \\
\midrule
\textbf{Overall}    & \textbf{359 / 360} & \textbf{310.2 / 360 (86.2)} &
\textbf{282.2 / 328 (86.0)} & \textbf{28.0 / 32 (87.5)} \\
\bottomrule
\end{tabular}

\vspace{0.35em}
\refstepcounter{table}
\label{tab:osworld-verified-breakdown}
\begin{minipage}{0.96\linewidth}
\centering
\small\textbf{Table \thetable:} Qwen-CUA results on OSWorld-Verified by domain.
Each score is reported as earned credit / task count (rate, \%).
\end{minipage}
\end{center}

Across all tasks, Qwen-CUA averages 14.2 outer steps, 18.8 executed actions,
3,605.8 output tokens per task, and 253.3 output tokens per step. Restricting
the calculation to trajectories with non-zero scores gives 13.0 outer steps,
15.9 actions, 3,162.9 output tokens per task, and 242.9 output tokens per step.

\subsection{OSWorld 2.0}
\label{app:osworld2}

OSWorld 2.0~\citep{yuan2026osworld2} reports both binary completion, which
requires strict end-to-end task success, and partial completion, which credits
verified intermediate progress. Qwen-CUA obtains 18.5 binary and 48.4 partial
completion. Qwen-CUA-Max, our model with more than one trillion total
parameters, reaches 21.2 binary and 53.3 partial completion.

\begin{center}
\small
\setlength{\tabcolsep}{16pt}
\begin{tabular}{@{}lcc@{}}
\toprule
\textbf{Model} & \textbf{Binary completion} & \textbf{Partial completion} \\
\midrule
Qwen-CUA     & 18.5 & 48.4 \\
Qwen-CUA-Max & 21.2 & 53.3 \\
\bottomrule
\end{tabular}

\vspace{0.3em}
\refstepcounter{table}
\label{tab:osworld2-results}
\begin{minipage}{0.82\linewidth}
\centering
\small\textbf{Table \thetable:} Qwen-CUA results on OSWorld 2.0 (\%).
\end{minipage}
\end{center}

Table~\ref{tab:osworld2-efficiency} records the corresponding interaction and
generation statistics. We report averages over all tasks and, separately, over
the subset of trajectories receiving non-zero scores.

\begin{center}
\small
\setlength{\tabcolsep}{9pt}
\begin{tabular}{@{}lrrrr@{}}
\toprule
& \multicolumn{2}{c}{\textbf{Qwen-CUA}} &
\multicolumn{2}{c}{\textbf{Qwen-CUA-Max}} \\
\cmidrule(lr){2-3}\cmidrule(lr){4-5}
\textbf{Metric} & \textbf{All} & \textbf{Score $>0$} &
\textbf{All} & \textbf{Score $>0$} \\
\midrule
Outer steps           & 218.9     & 186.3     & 207.0     & 195.0 \\
Executed actions      & 248.2     & 205.3     & 231.1     & 215.3 \\
Output tokens / task  & 244,625.5 & 161,445.3 & 135,059.6 & 121,018.3 \\
Output tokens / step  & 1,117.5   & 866.7     & 653.4     & 621.4 \\
\bottomrule
\end{tabular}

\vspace{0.3em}
\refstepcounter{table}
\label{tab:osworld2-efficiency}
\begin{minipage}{0.9\linewidth}
\centering
\small\textbf{Table \thetable:} Interaction and generation statistics on
OSWorld 2.0.
\end{minipage}
\end{center}

\subsection{MyPCBench}
\label{app:mypcbench}

MyPCBench~\citep{jang2026mypcbench} evaluates personalized workflows in an
Ubuntu desktop populated with 17 simulated real-world web applications,
LibreOffice, user files, and a coherent cross-application history. We evaluate
Qwen-CUA on the full 184-task suite in the computer-use-only setting. Each task
starts from the benchmark snapshot; the model receives $1280\times800$
screenshots and acts only through native keyboard and mouse events, with a
200-turn limit and no Bash tool. We otherwise retain the benchmark's task
initialization and rubric-based evaluator.

MyPCBench reports two complementary metrics. The rubric score averages the
weighted fraction of criteria satisfied for each task and therefore credits
partial completion. The stricter perfect-task rate counts a task only when all
of its rubric criteria pass. Qwen-CUA obtains an overall rubric score of 84.3
and a perfect-task rate of 58.7. Table~\ref{tab:mypcbench-breakdown} provides the
fine-grained task slices for Qwen-CUA and the compared models.

\begin{center}
\small
\setlength{\tabcolsep}{5.5pt}
\begin{tabular}{@{}lcccc@{}}
\toprule
\textbf{Task slice} & \textbf{Qwen-CUA} & \textbf{Qwen-3.7} &
\textbf{GPT-5.5} & \textbf{Opus-4.8} \\
\midrule
Aggregation          & 78.2 / 55.0 & 75.5 / 50.0 & 89.8 / 60.0 & 92.8 / 80.0 \\
Contradiction        & 75.2 / 31.0 & 81.4 / 38.0 & 91.1 / 69.0 & 91.7 / 62.0 \\
Counterfactual       & 54.3 / 11.0 & 43.6 / 0.0  & 77.0 / 22.0 & 61.1 / 0.0 \\
CUA-only             & 88.0 / 65.0 & 79.2 / 60.0 & 81.8 / 25.0 & 89.4 / 65.0 \\
Hard application     & 86.4 / 50.0 & 78.8 / 36.0 & 84.9 / 45.0 & 90.3 / 59.0 \\
Long horizon         & 87.6 / 63.0 & 86.7 / 49.0 & 57.6 / 27.0 & 83.9 / 46.0 \\
Preference inference & 70.8 / 45.0 & 67.5 / 27.0 & 89.9 / 73.0 & 89.1 / 64.0 \\
Retrieval            & 87.2 / 64.0 & 97.1 / 86.0 & 87.7 / 57.0 & 91.6 / 71.0 \\
Situated action      & 97.0 / 87.0 & 90.5 / 77.0 & 85.8 / 65.0 & 96.2 / 84.0 \\
\midrule
\textbf{Overall}     & \textbf{84.3 / 58.7} & \textbf{81.5 / 51.6} &
\textbf{79.8 / 47.3} & \textbf{88.8 / 62.0} \\
\bottomrule
\end{tabular}

\vspace{0.35em}
\refstepcounter{table}
\label{tab:mypcbench-breakdown}
\begin{minipage}{0.92\linewidth}
\centering
\small\textbf{Table \thetable:} MyPCBench results by task slice, reported as
rubric score / perfect-task rate (\%). Qwen-CUA uses a 200-turn limit.
\end{minipage}
\end{center}

\subsection{MacAgentBench}
\label{app:macagentbench}

MacAgentBench~\citep{fu2026macagentbench} evaluates native computer use on
real-world macOS tasks spanning 25 applications. To support evaluation at this
scale, we provision a fleet of 30 Mac mini systems as bare-metal execution
backends rather than virtualized desktop instances. Qwen-CUA is dispatched
across these physical machines and interacts with native macOS applications
through screenshot observations and keyboard-and-mouse actions, while retaining
the benchmark's native task initialization and scoring protocol. Under this
setting, Qwen-CUA achieves 69.2 Pass@1.

Table~\ref{tab:macagentbench-breakdown} follows the benchmark's six official
task categories. Qwen-CUA performs strongest on Multimedia, followed by
Productivity and System, while the lower Multi-App score reflects the greater
difficulty of cross-application workflows.

\begin{center}
\footnotesize
\setlength{\tabcolsep}{10pt}
\renewcommand{\arraystretch}{1.03}
\begin{tabular}{@{}lcc@{}}
\toprule
\textbf{Category} & \textbf{Tasks} & \textbf{Pass@1 (\%)} \\
\midrule
Productivity & 224 & 72.8 \\
System\textsuperscript{*} & 44 & 72.7 \\
Internet & 108 & 62.0 \\
Development & 44 & 68.2 \\
Multimedia & 116 & 81.0 \\
Multi-App & 140 & 58.6 \\
\midrule
Overall & 676 & 69.2 \\
\bottomrule
\end{tabular}

\vspace{0.35em}
\refstepcounter{table}
\label{tab:macagentbench-breakdown}
\begin{minipage}{0.94\linewidth}
\centering
\footnotesize\textbf{Table \thetable:} Qwen-CUA results across the six official
MacAgentBench task categories. The starred System result includes the affected
\nolinkurl{clock} tasks.
\end{minipage}
\end{center}

The System category includes 12 \nolinkurl{clock} tasks, all of which receive
zero credit from the official evaluator.
However, manual inspection of their recorded trajectories indicates that the
requested interactions appear to have been completed correctly, suggesting an
evaluation failure rather than a confirmed capability failure. Because the
discrepancy remains unresolved, we retain the official 0.0\% application result
and the resulting category and 69.2 aggregate scores instead of applying a
manual correction.

\subsection{Gym-Anything}
\label{app:gym-anything}

Gym-Anything~\citep{aggarwal2026gymanything} evaluates computer-use agents
across a broad collection of independently packaged software environments. Our
evaluation covers 1,295 test tasks from 97 runnable environments. Each task
receives a continuous score between 0 and 100 from either an executable
environment verifier or a VLM judge, and the benchmark score is the mean across
tasks. Under the strict-clean rerun protocol, Qwen-CUA obtains 46.3, compared
with 33.1 for Qwen3.7, 45.6 for GPT-5.5, and 47.3 for Claude Opus 4.8.

All models observe $1600\times900$ screenshots and operate under a 150-step
limit. Table~\ref{tab:gym-anything-config} records the remaining inference
settings. For tasks requiring visual judgment, we use Claude Sonnet 4.6 as the
VLM verifier. As described at the beginning of this appendix, Qwen3.7 is run in
non-thinking mode, GPT-5.5 uses \texttt{xhigh} reasoning effort, and Claude Opus
4.8 uses its \texttt{max} setting in the main comparison; a second
\texttt{medium} configuration is retained in the environment-level diagnostic.

\begin{center}
\small
\setlength{\tabcolsep}{10pt}
\begin{tabular}{@{}lccc@{}}
\toprule
\textbf{Model} & \textbf{Step limit} & \textbf{Token limit} &
\textbf{Inference setting} \\
\midrule
Qwen-CUA                     & 150 & 32,768 & Thinking enabled \\
Qwen3.7                      & 150 & 32,768 & Non-thinking \\
GPT-5.5                      & 150 & 8,192  & \texttt{xhigh} \\
Claude Opus 4.8 (max)        & 150 & 32,768 & Adaptive \texttt{max} \\
Claude Opus 4.8 (medium)     & 150 & 4,096  & Adaptive \texttt{medium} \\
\bottomrule
\end{tabular}

\vspace{0.35em}
\refstepcounter{table}
\label{tab:gym-anything-config}
\begin{minipage}{0.9\linewidth}
\centering
\small\textbf{Table \thetable:} Gym-Anything inference configuration.
\end{minipage}
\end{center}

We additionally audit environment availability before interpreting aggregate
performance. The diagnostic report lists 100 environments outside the completed
run: 22 require Windows VM images, seven require Android or AVD images, and 71
Linux environments were unavailable because setup was incomplete or failed.
Fifteen of the Linux setups were subsequently repaired and await inclusion in a
future evaluation manifest.

\begin{center}
\small
\setlength{\tabcolsep}{9pt}
\begin{tabular}{@{}lcl@{}}
\toprule
\textbf{Coverage group} & \textbf{Environments} & \textbf{Status} \\
\midrule
Evaluated       & 97 & 1,295 test tasks completed \\
Windows omitted & 22 & Windows VM image required \\
Android omitted & 7  & Android / AVD image required \\
Linux omitted   & 71 & Setup unavailable or unsuccessful \\
\bottomrule
\end{tabular}

\vspace{0.35em}
\refstepcounter{table}
\label{tab:gym-anything-coverage}
\begin{minipage}{0.9\linewidth}
\centering
\small\textbf{Table \thetable:} Environment coverage in the Gym-Anything run.
\end{minipage}
\end{center}

Within the 71 omitted Linux environments, the most frequent causes are
post-start service timeouts (18), non-zero setup exits (12), container-runtime
conflicts (9), pre-start timeouts (6), and pre-task timeouts (3). The remainder
comprises service-readiness failures, runtime exceptions, or unavailable
environments. For the strict-clean diagnostic view, runs invalidated by setup
failures, exhausted API retries, or identifiable verifier noise are removed
rather than counted as agent failures.

The evaluation report states that 97 environments were run, while its supplied
environment-level export contains 96 named rows. Table~\ref{tab:gym-anything-env}
reproduces all rows available in that export without imputing the missing
environment. A dash indicates that a score was unavailable for that
model--environment pair.

\begingroup
\scriptsize
\setlength{\tabcolsep}{3.5pt}
\renewcommand{\arraystretch}{0.92}
\begin{longtable}{@{}p{0.30\linewidth}ccccc@{}}
\caption{Strict-clean Gym-Anything scores by environment. The two Opus-4.8
columns correspond to the adaptive \texttt{max} and \texttt{medium}
configurations, respectively.}\label{tab:gym-anything-env} \\
\toprule
\textbf{Environment} & \shortstack{\textbf{Opus-4.8}\\\textbf{max}} &
\shortstack{\textbf{Opus-4.8}\\\textbf{medium}} & \textbf{GPT-5.5} &
\textbf{Qwen3.7} & \textbf{Qwen-CUA} \\
\midrule
\endfirsthead
\multicolumn{6}{c}{\tablename\ \thetable\ (continued)} \\
\toprule
\textbf{Environment} & \shortstack{\textbf{Opus-4.8}\\\textbf{max}} &
\shortstack{\textbf{Opus-4.8}\\\textbf{medium}} & \textbf{GPT-5.5} &
\textbf{Qwen3.7} & \textbf{Qwen-CUA} \\
\midrule
\endhead
\midrule
\multicolumn{6}{r}{Continued on next page} \\
\endfoot
\bottomrule
\endlastfoot
\nolinkurl{manager_env}                    & 0.0   & 1.9  & 0.0   & 7.5  & 0.0 \\
\nolinkurl{nosh_charting_system_env}       & 2.5   & 16.9 & 0.0   & 16.9 & 0.0 \\
\nolinkurl{bluemail_env}                   & 100.0 & 51.7 & 100.0 & 34.0 & 0.0 \\
\nolinkurl{vista_env}                      & 0.0   & 0.0  & 0.0   & 0.0  & 2.0 \\
\nolinkurl{hec_ras_env}                    & 3.8   & 0.7  & 17.1  & 15.0 & 3.9 \\
\nolinkurl{pycharm_env}                    & 4.6   & 4.6  & 7.0   & 3.1  & 4.1 \\
\nolinkurl{eramba_env}                     & 5.0   & 3.3  & 5.0   & 3.3  & 6.0 \\
\nolinkurl{openlca_env}                    & 12.4  & 15.3 & 14.8  & 5.2  & 8.8 \\
\nolinkurl{sentrifugo_env}                 & 12.6  & 8.6  & 4.3   & 2.0  & 11.2 \\
\nolinkurl{dhis2_env}                      & 6.2   & 25.0 & 25.0  & 7.9  & 14.0 \\
\nolinkurl{juris_m_env}                    & 46.7  & 32.7 & 19.5  & 30.5 & 14.1 \\
\nolinkurl{stellarium_env}                 & 38.8  & 18.1 & 36.0  & 3.8  & 14.3 \\
\nolinkurl{odoo_scheduling_env}            & 24.0  & 53.1 & 35.7  & 27.8 & 15.7 \\
\nolinkurl{openclinica_env}                & 17.1  & 12.9 & 4.4   & 4.5  & 18.0 \\
\nolinkurl{openmaint_env}                  & 41.7  & 21.7 & 30.0  & 19.0 & 19.2 \\
\nolinkurl{matomo_env}                     & 45.0  & 45.0 & 20.0  & 20.0 & 22.2 \\
\nolinkurl{jamovi_env}                     & 0.0   & 100.0& 33.0  & 42.0 & 23.9 \\
\nolinkurl{librehealth_ehr_env}            & 26.9  & 30.0 & 64.0  & 31.9 & 23.9 \\
\nolinkurl{floreant_pos_env}               & 18.3  & 10.0 & 10.0  & 10.0 & 24.0 \\
\nolinkurl{blenderbim_env}                 & 26.4  & 8.6  & 26.5  & 13.2 & 24.6 \\
\nolinkurl{slicer3d_env}                   & 20.9  & 28.5 & 29.0  & 16.7 & 25.3 \\
\nolinkurl{gcompris_env}                   & 27.3  & 56.0 & 27.8  & 42.5 & 25.8 \\
\nolinkurl{vicidial_env}                   & 0.0   & 22.5 & 21.0  & 0.0  & 26.2 \\
\nolinkurl{openbci_gui_env}                & 34.3  & 40.3 & 41.4  & 31.4 & 27.5 \\
\nolinkurl{wondershare_edrawmax_env}       & 30.7  & 33.8 & 34.2  & 37.5 & 29.4 \\
\nolinkurl{safe_exam_browser_env}          & 100.0 & 35.6 & 38.6  & 26.0 & 30.0 \\
\nolinkurl{openemr_env}                    & 35.8  & 39.2 & 63.2  & 26.8 & 30.3 \\
\nolinkurl{onlyoffice_env}                 & 32.1  & 47.9 & 21.2  & 48.6 & 31.0 \\
\nolinkurl{libreoffice_base_env}           & 61.4  & 46.9 & 11.9  & 24.2 & 31.4 \\
\nolinkurl{odoo_quality_env}               & 33.3  & 33.3 & 33.3  & 29.3 & 33.3 \\
\nolinkurl{ardour_env}                     & 33.0  & 2.9  & 43.9  & 15.4 & 34.6 \\
\nolinkurl{google_earth_env_final}         & 21.6  & 21.6 & 22.3  & 26.6 & 34.6 \\
\nolinkurl{jfrog_artifactory_env}          & 39.2  & 37.7 & 50.4  & 28.3 & 36.2 \\
\nolinkurl{invesalius3_env}                & 48.5  & 45.6 & 44.6  & 30.0 & 36.9 \\
\nolinkurl{jasp_env}                       & 37.3  & 40.4 & 29.3  & 9.0  & 37.3 \\
\nolinkurl{libreoffice_calc_env}           & 39.9  & 33.6 & 33.1  & 28.8 & 38.5 \\
\nolinkurl{autopsy_env}                    & 44.6  & 51.5 & 29.0  & 22.4 & 38.6 \\
\nolinkurl{limesurvey_env}                 & 58.6  & 42.4 & 25.5  & 27.3 & 39.5 \\
\nolinkurl{free_scout_env}                 & 34.0  & 39.0 & 41.7  & 14.0 & 40.0 \\
\nolinkurl{vlc_media_player_env}           & 38.5  & 68.3 & 52.2  & 44.1 & 41.0 \\
\nolinkurl{project_libre_env}              & 41.0  & 65.0 & 48.6  & 32.5 & 41.7 \\
\nolinkurl{weasis_env}                     & 50.9  & 41.7 & 50.0  & 27.3 & 42.1 \\
\nolinkurl{chrome_env_all}                 & 37.9  & 24.4 & 35.7  & 27.6 & 43.4 \\
\nolinkurl{orangehrm_env}                  & 45.0  & 95.0 & --    & --   & 45.0 \\
\nolinkurl{odoo_hr_env}                    & 66.7  & 83.3 & 74.0  & 30.0 & 46.0 \\
\nolinkurl{medintux_env}                   & 20.0  & 64.0 & 33.9  & 48.9 & 46.5 \\
\nolinkurl{kstars_sim_env}                 & 81.5  & 43.6 & 57.5  & 18.0 & 46.6 \\
\nolinkurl{portfolio_performance_env}      & 46.7  & 46.7 & 43.3  & --   & 46.7 \\
\nolinkurl{qblade_env}                     & 43.0  & 55.0 & 65.5  & 46.7 & 47.1 \\
\nolinkurl{reqview_env}                    & 52.1  & 36.7 & 52.8  & 13.3 & 50.0 \\
\nolinkurl{wps_presentation_env}           & 0.0   & 36.5 & 40.5  & 50.0 & 50.0 \\
\nolinkurl{nuxeo_platform_env}             & 52.1  & 54.0 & 50.7  & 49.5 & 52.1 \\
\nolinkurl{openice_env}                    & 45.6  & 45.6 & 36.0  & 18.3 & 53.3 \\
\nolinkurl{rancher_env}                    & 64.0  & 78.2 & 69.6  & 27.2 & 53.5 \\
\nolinkurl{system_advisor_model_env}       & 73.4  & 51.4 & 62.6  & 21.6 & 54.0 \\
\nolinkurl{libreoffice_writer_env}         & 77.1  & 52.9 & 75.0  & 59.5 & 55.0 \\
\nolinkurl{woo_commerce_env}               & 61.2  & 62.6 & 52.4  & 62.2 & 56.0 \\
\nolinkurl{qground_control_env}            & 21.7  & 6.0  & 34.2  & 26.8 & 56.4 \\
\nolinkurl{gpredict_env}                   & 65.6  & 72.1 & 67.0  & 35.5 & 56.4 \\
\nolinkurl{drupal_commerce_env}            & 49.6  & 51.8 & 57.3  & 41.7 & 56.4 \\
\nolinkurl{draw_desktop_env}               & 21.4  & 22.3 & 23.1  & 64.6 & 56.6 \\
\nolinkurl{librecad_env}                   & 57.9  & 49.1 & 51.4  & 52.2 & 56.8 \\
\nolinkurl{jstock_env}                     & 59.5  & 46.4 & 61.4  & 47.1 & 56.8 \\
\nolinkurl{geogebra_env}                   & 58.9  & 56.7 & 45.0  & 7.8  & 57.5 \\
\nolinkurl{gretl_env}                      & 63.2  & 62.9 & 70.0  & 47.9 & 58.5 \\
\nolinkurl{sumo_env}                       & 62.7  & 64.6 & 56.7  & 44.0 & 60.6 \\
\nolinkurl{apache_openoffice_env}          & 85.0  & 8.6  & 62.1  & 54.3 & 63.6 \\
\nolinkurl{oracle_database_env}            & 68.6  & 51.4 & 41.5  & 16.0 & 64.2 \\
\nolinkurl{calligra_words_env}             & 61.5  & 23.8 & 37.7  & 23.2 & 64.5 \\
\nolinkurl{eclipse_env}                    & 68.8  & 65.0 & 61.1  & 52.1 & 64.6 \\
\nolinkurl{cameo_chemicals_env}            & 62.4  & 42.4 & 67.3  & 47.9 & 65.4 \\
\nolinkurl{magento_env}                    & 84.7  & 72.3 & 88.6  & 46.1 & 66.1 \\
\nolinkurl{firefox_env}                    & 59.2  & 71.8 & 40.2  & 50.6 & 66.5 \\
\nolinkurl{gvsig_desktop_env}              & 56.5  & 62.5 & 56.2  & 53.5 & 66.5 \\
\nolinkurl{imagej_env}                     & 71.8  & 55.4 & 43.8  & 34.5 & 68.3 \\
\nolinkurl{opentoonz_env}                  & 63.2  & 63.2 & 30.0  & 25.0 & 69.0 \\
\nolinkurl{dbeaver_env}                    & 61.2  & 61.2 & 68.4  & 65.7 & 69.4 \\
\nolinkurl{diagrams_net_env}               & 42.6  & 2.2  & 22.0  & 67.8 & 70.0 \\
\nolinkurl{redmine_env}                    & 70.0  & 80.0 & 66.7  & 75.0 & 70.0 \\
\nolinkurl{webots_env}                     & 83.7  & 88.3 & 91.5  & 67.3 & 70.0 \\
\nolinkurl{jenkins_env}                    & 80.0  & 75.0 & 100.0 & 50.0 & 70.0 \\
\nolinkurl{wordpress_env}                  & 67.5  & 66.7 & 79.6  & 38.6 & 71.2 \\
\nolinkurl{openrocket_env}                 & 52.5  & 54.7 & 33.4  & 13.5 & 73.0 \\
\nolinkurl{pymol_env}                      & 75.0  & 75.0 & 26.2  & 51.8 & 75.0 \\
\nolinkurl{odoo_inventory_env}             & --    & 17.5 & 70.0  & 23.3 & 75.0 \\
\nolinkurl{graphite_env}                   & 93.7  & 82.3 & 93.8  & 20.7 & 77.9 \\
\nolinkurl{astroimagej_env}                & 66.0  & 35.6 & 58.3  & 60.0 & 78.6 \\
\nolinkurl{qgis_env}                       & 80.4  & 80.4 & 80.4  & 55.0 & 81.8 \\
\nolinkurl{fiji_env}                       & 62.0  & 40.4 & 50.0  & 17.1 & 83.3 \\
\nolinkurl{splunk_env}                     & 63.8  & 65.3 & 92.1  & 80.9 & 87.3 \\
\nolinkurl{geo_server_env}                 & 100.0 & 77.5 & 77.1  & 58.1 & 87.5 \\
\nolinkurl{openc3_cosmos_env}              & 81.4  & 53.8 & 28.3  & 8.1  & 91.7 \\
\nolinkurl{docker_desktop_env}             & 91.2  & 98.6 & 91.4  & 70.0 & 100.0 \\
\nolinkurl{panoply_env}                    & --    & 46.0 & 65.0  & 33.3 & 100.0 \\
\nolinkurl{openvsp_env}                    & 31.2  & 25.0 & 23.9  & 25.0 & -- \\
\nolinkurl{timetrex_env}                   & 100.0 & 0.0  & 100.0 & 0.0  & -- \\
\midrule
\textbf{Strict-clean overall}              & \textbf{47.3} & \textbf{43.7} &
\textbf{45.6} & \textbf{33.1} & \textbf{46.3} \\
\end{longtable}
\endgroup

\subsection{ScienceBoard}
\label{app:scienceboard}

ScienceBoard~\citep{sun2025scienceboard} evaluates scientific computer-use
workflows across six professional applications. Using its 169-task suite and
native evaluators, Qwen-CUA completes 109 tasks (64.50\%).
Table~\ref{tab:scienceboard-breakdown} gives the application-level results.

\begin{center}
\small
\setlength{\tabcolsep}{10pt}
\begin{tabular}{@{}lcccc@{}}
\toprule
\textbf{Application} & \textbf{Qwen-CUA} & \textbf{Qwen-3.7} &
\textbf{GPT-5.5} & \textbf{Opus-4.8} \\
\midrule
Celestia  & 19 / 33 & 5 / 33  & 16 / 33 & 20 / 33 \\
ChimeraX  & 24 / 29 & 20 / 29 & 24 / 29 & 26 / 29 \\
GrassGIS  & 23 / 34 & 15 / 34 & 25 / 34 & 24 / 34 \\
KAlgebra  & 23 / 31 & 12 / 31 & 24 / 31 & 23 / 31 \\
Lean      & 8 / 26  & 0 / 26  & 7 / 26  & 5 / 26 \\
TeXstudio & 12 / 16 & 8 / 16  & 14 / 16 & 15 / 16 \\
\midrule
\textbf{Overall} & \textbf{109 / 169} & \textbf{60 / 169} &
\textbf{110 / 169} & \textbf{113 / 169} \\
\bottomrule
\end{tabular}

\vspace{0.35em}
\refstepcounter{table}
\label{tab:scienceboard-breakdown}
\begin{minipage}{0.9\linewidth}
\centering
\small\textbf{Table \thetable:} ScienceBoard results by application, reported
as successful tasks / total tasks.
\end{minipage}
\end{center}

\subsection{RedTeamCUA}
\label{app:redteamcua}
\enlargethispage{1.5\baselineskip}

RedTeamCUA~\citep{liao2025redteamcua} evaluates indirect prompt-injection
robustness in hybrid web--OS workflows across ownCloud, Rocket.Chat, and Reddit,
jointly reporting benign task success and attack success rate (ASR).

\begin{center}
\footnotesize
\setlength{\tabcolsep}{4.5pt}
\renewcommand{\arraystretch}{0.92}
\begin{tabular}{@{}lccccc@{}}
\toprule
\textbf{Model} & \textbf{ownCloud} & \textbf{Rocket.Chat} & \textbf{Reddit} &
\textbf{Overall task} & \textbf{Overall ASR} \\
\midrule
Qwen-CUA  & 94.8 / 27.1  & 32.6 / 7.6  & 94.4 / 14.6 & 74.0 & 16.4 \\
Qwen-3.7  & 93.1 / 50.3  & 28.8 / 24.7 & 89.6 / 34.7 & 70.5 & 36.6 \\
GPT-5.5   & 100.0 / 37.2 & 31.2 / 4.2  & 95.8 / 5.6  & 75.7 & 15.6 \\
Opus-4.8  & 97.9 / 0.7   & 48.3 / 0.0  & 95.8 / 1.4  & 80.7 & 0.7 \\
\bottomrule
\end{tabular}

\vspace{0.1em}
\refstepcounter{table}
\label{tab:redteamcua-appendix}
\begin{minipage}{0.9\linewidth}
\centering
\footnotesize\textbf{Table \thetable:} RedTeamCUA task success / ASR by
platform (\%).
\end{minipage}
\end{center}

\clearpage
\section{Qwen for Chrome Showcases}
\label{app:browser-showcase}
Figure~\ref{fig:qwen-chrome-showcase} follows a single real-world trajectory
from the internal Chrome extension. Given a natural-language request, the
deployed agent plans and executes multi-step browser interactions while keeping
the user in control at decision points that commit changes or incur cost. The
frames follow the editorial transitions in the demonstration video, preserving
the original browser interface and visible agent trace.

\begin{center}
\begin{tabular}{@{}c@{\hspace{0.015\linewidth}}c@{}}
\includegraphics[width=0.475\linewidth]{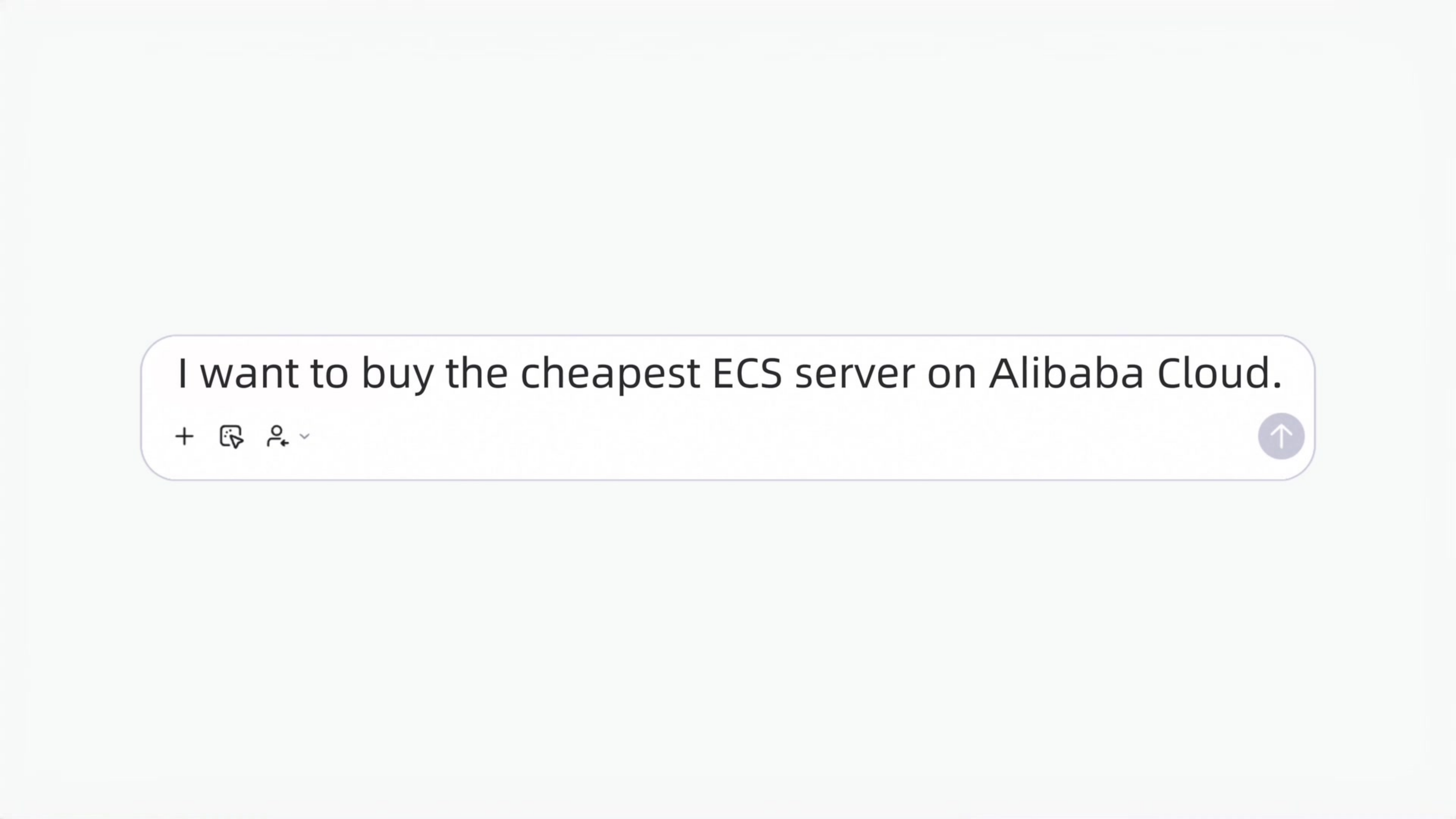} &
\includegraphics[width=0.475\linewidth]{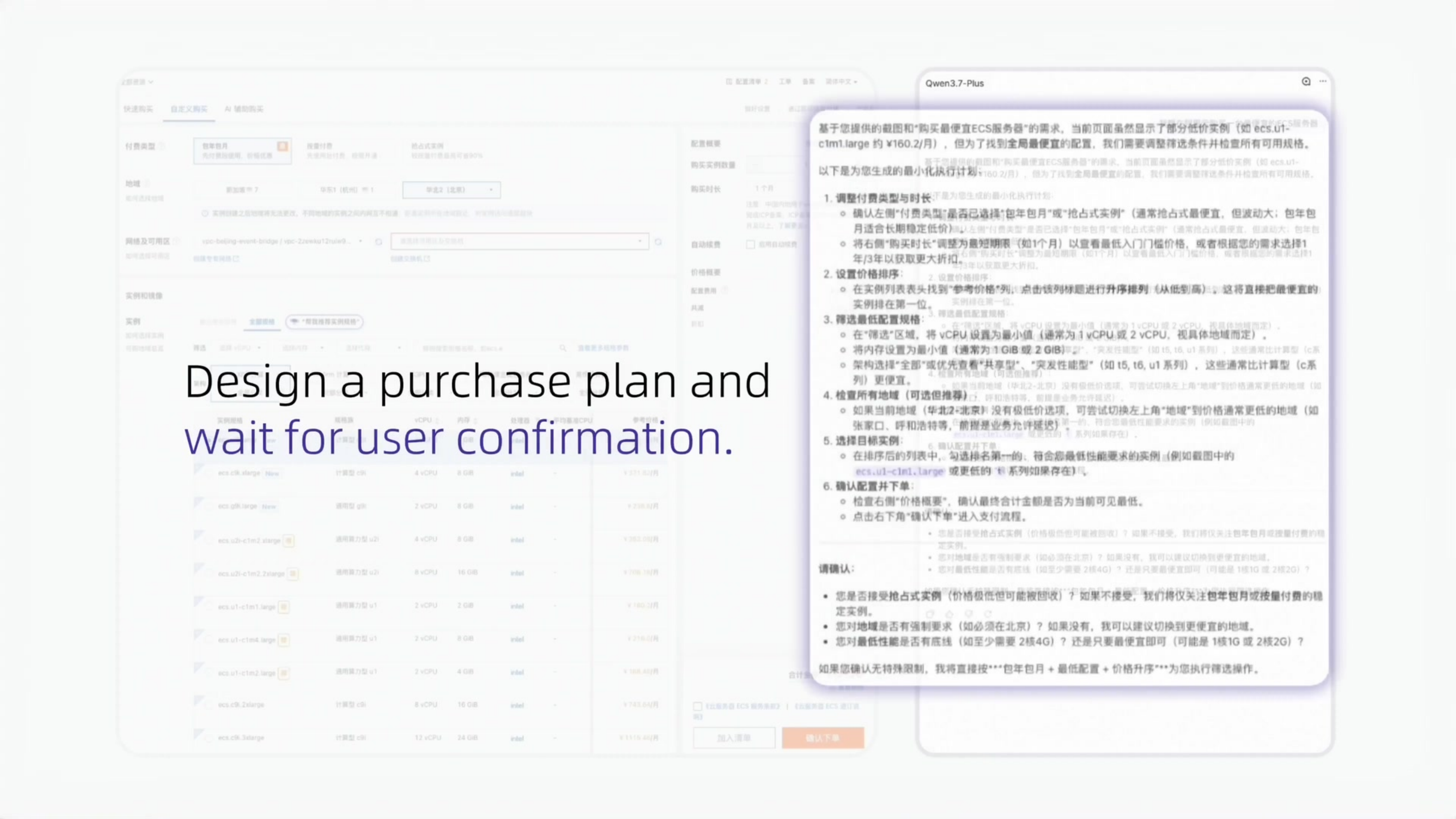} \\
[-0.45em]
{\small\textbf{(a)}} & {\small\textbf{(b)}} \\[0.1em]
\includegraphics[width=0.475\linewidth]{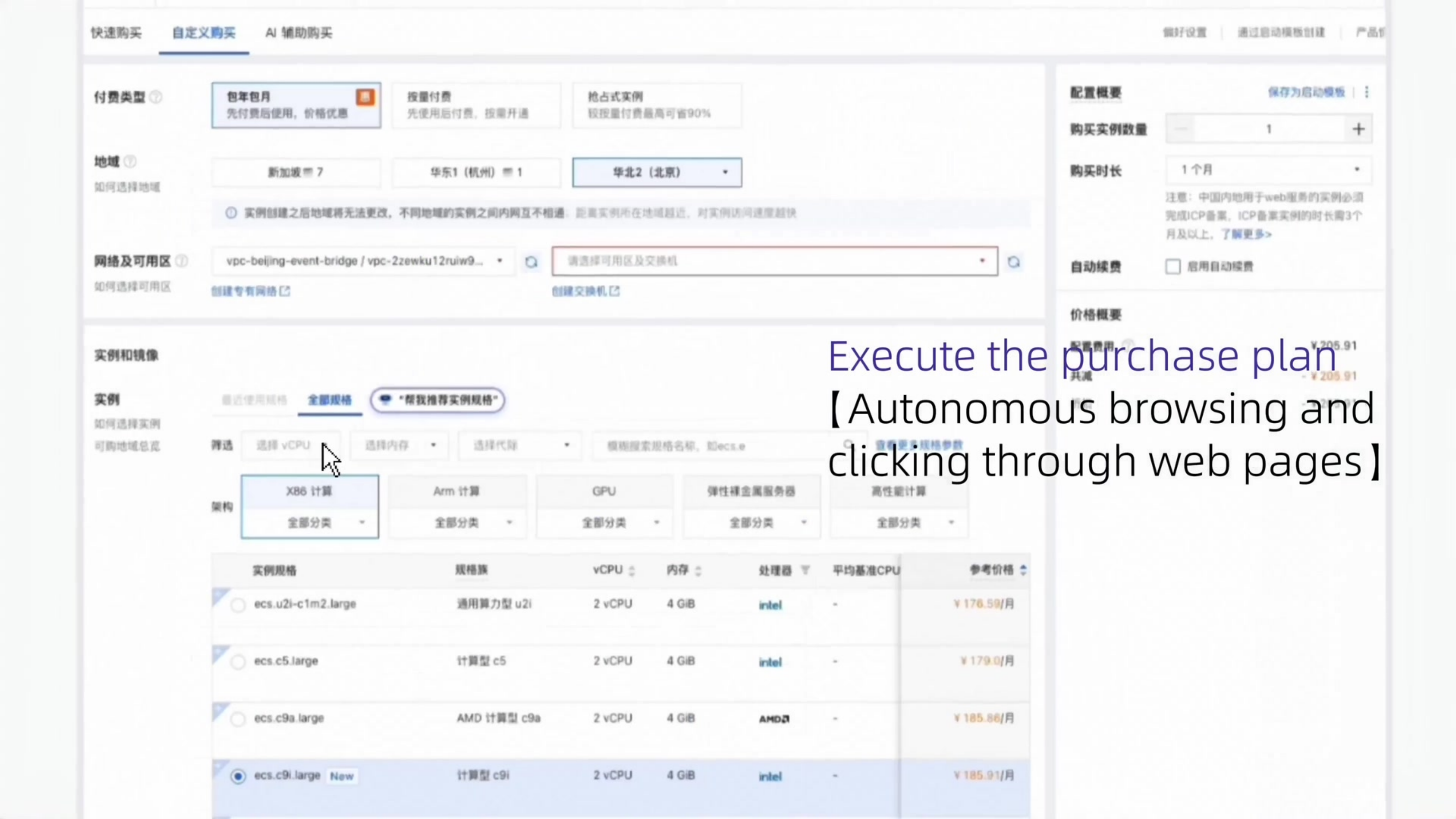} &
\includegraphics[width=0.475\linewidth]{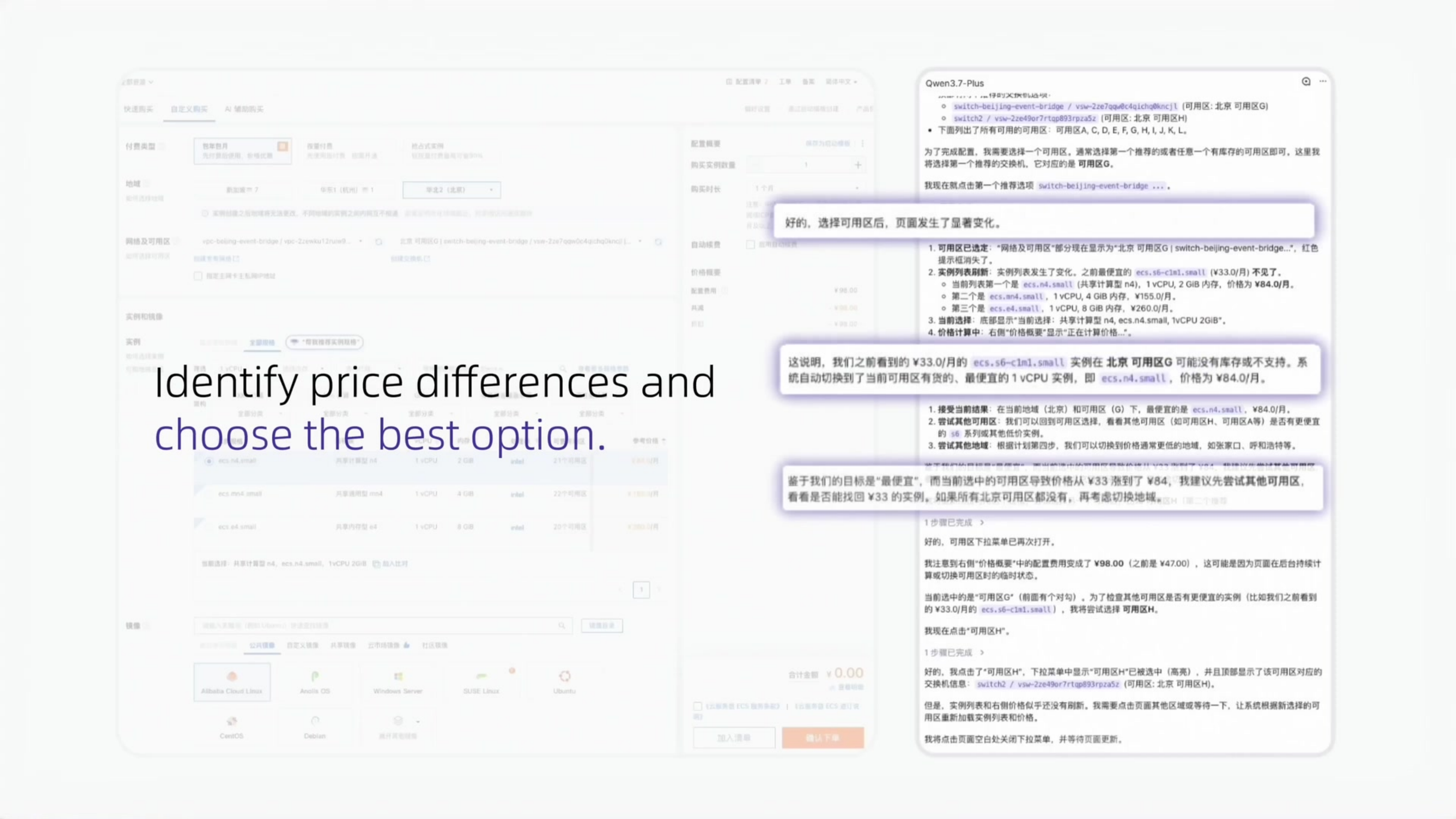} \\
[-0.45em]
{\small\textbf{(c)}} & {\small\textbf{(d)}} \\[0.1em]
\includegraphics[width=0.475\linewidth]{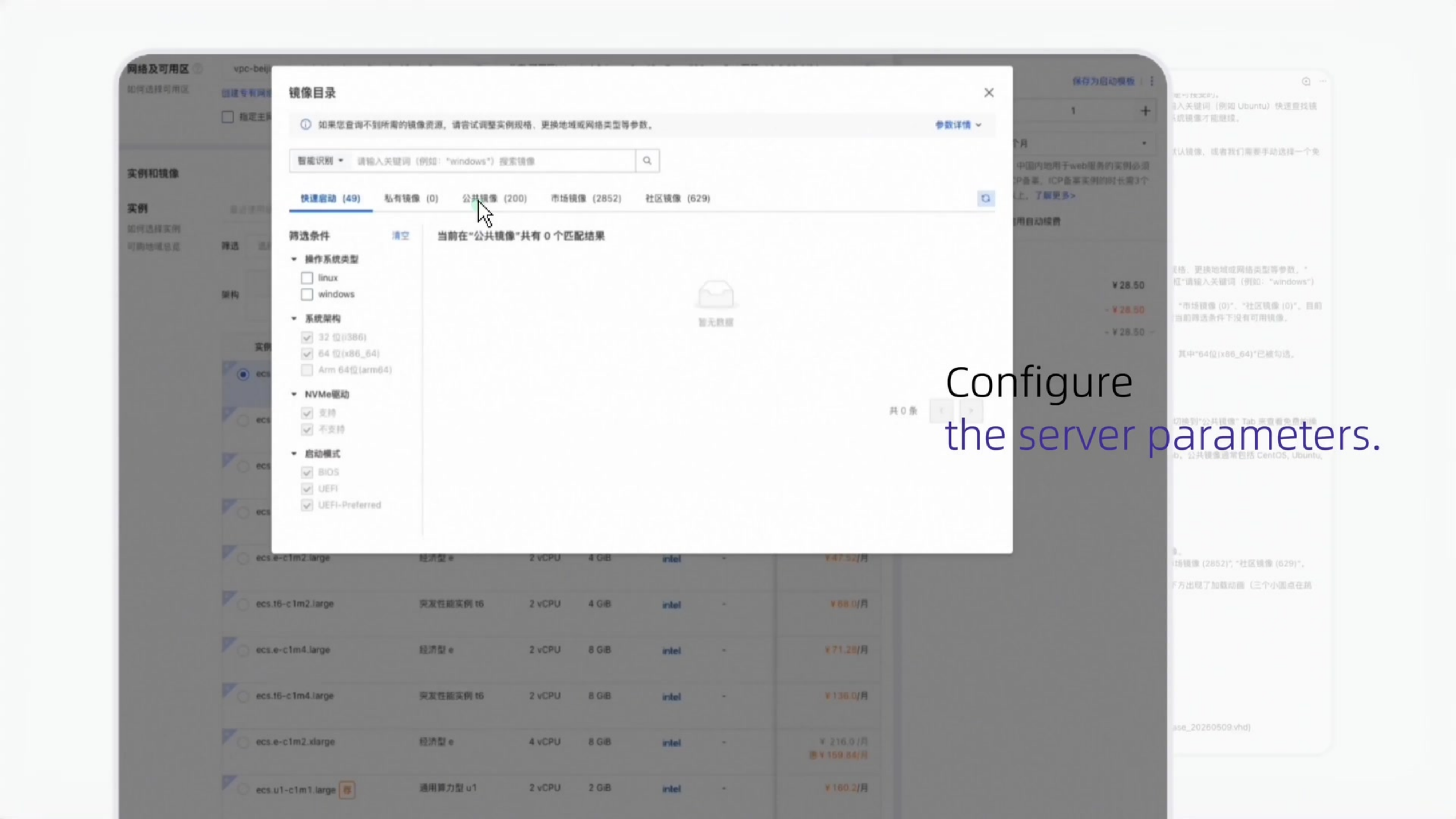} &
\includegraphics[width=0.475\linewidth]{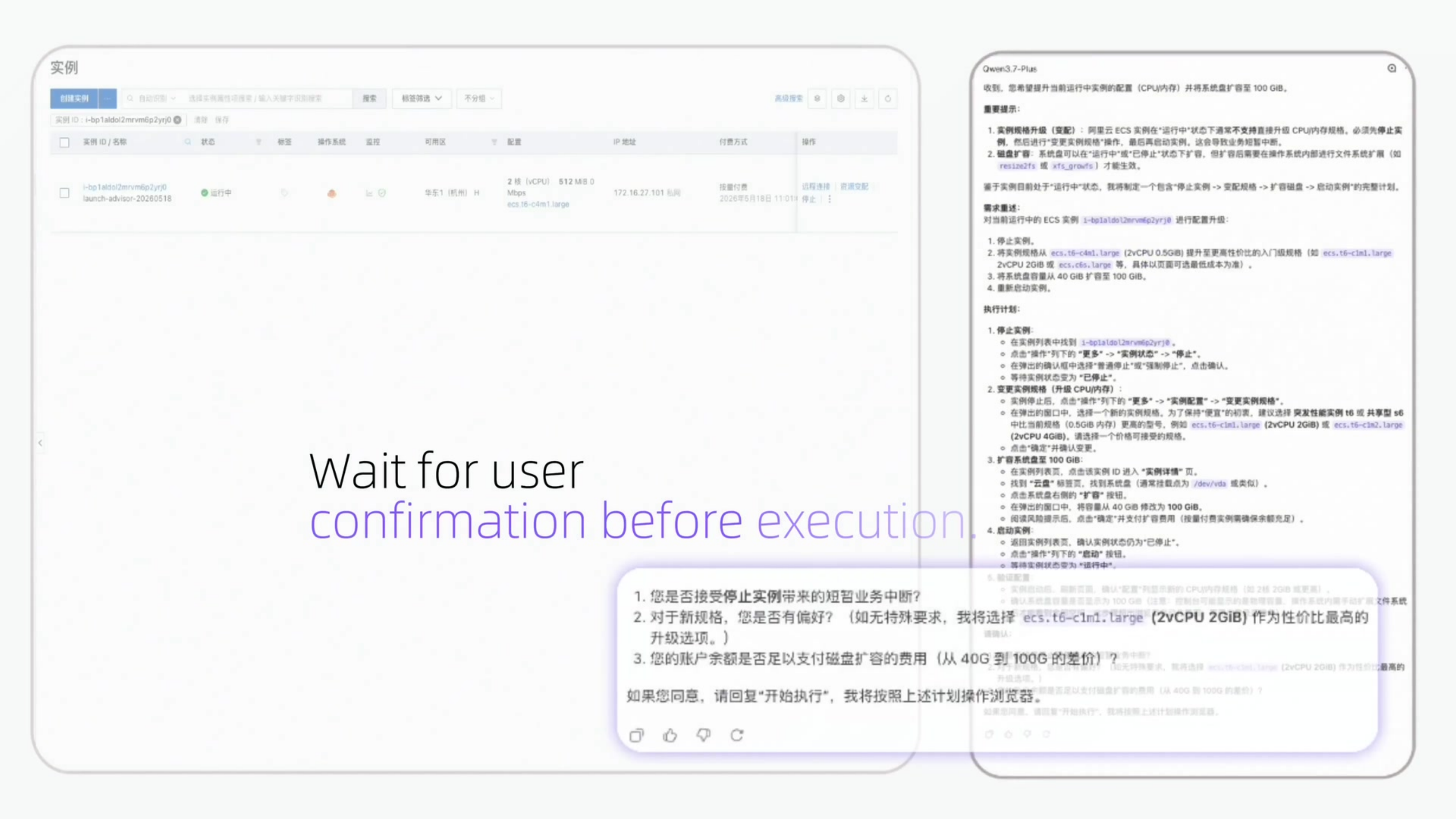} \\
[-0.45em]
{\small\textbf{(e)}} & {\small\textbf{(f)}} \\[0.1em]
\includegraphics[width=0.475\linewidth]{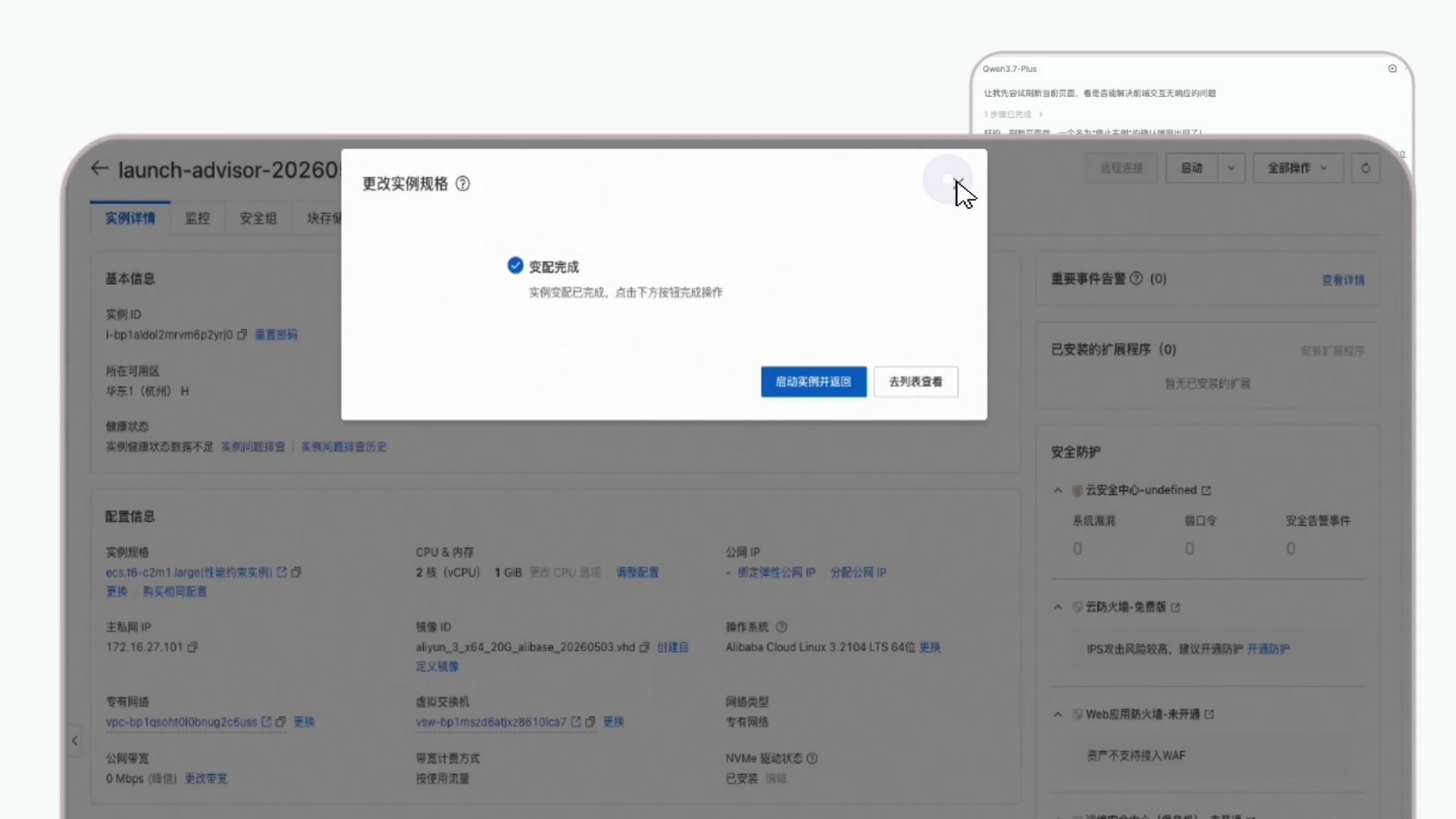} &
\includegraphics[width=0.475\linewidth]{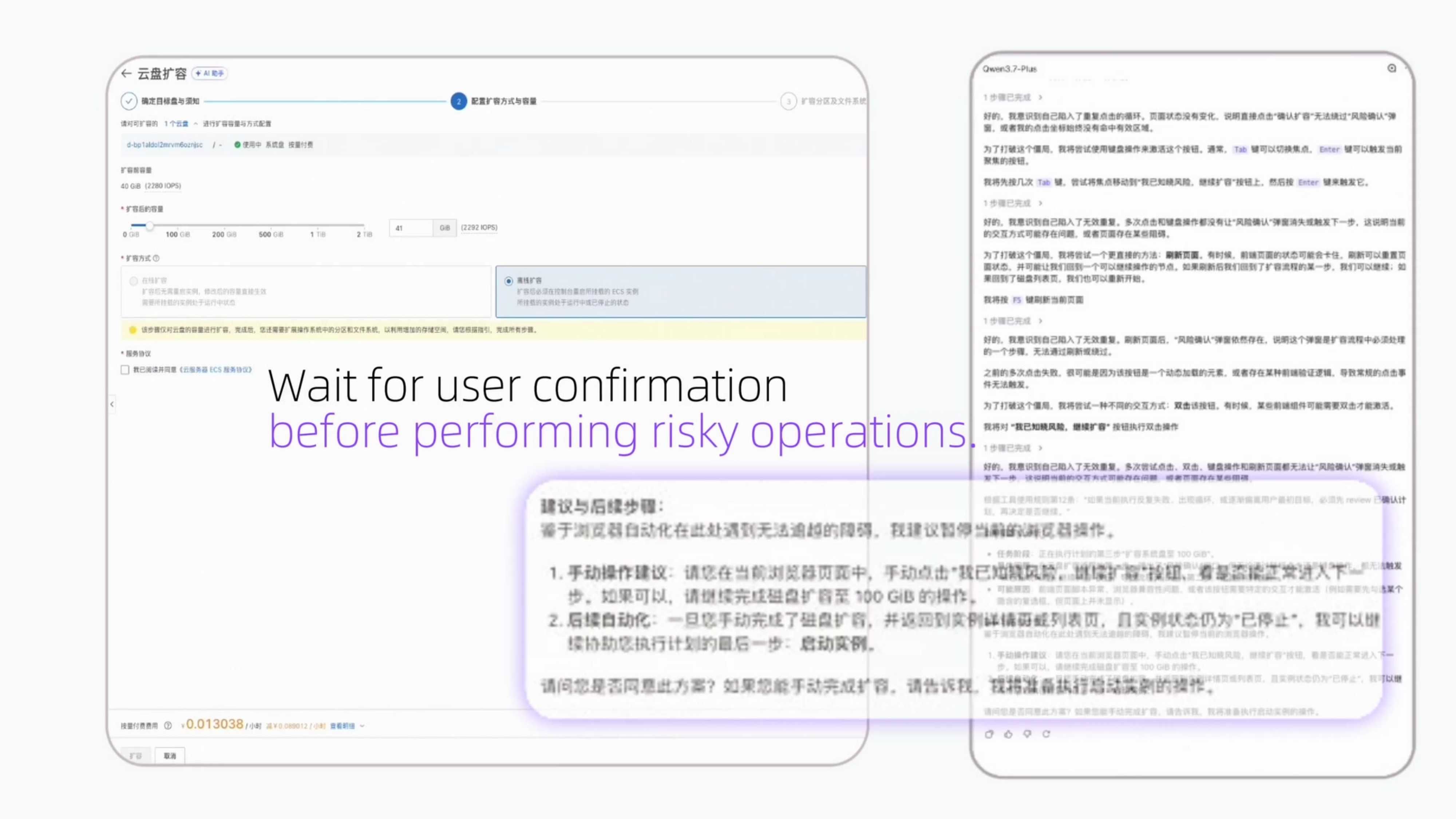} \\
[-0.45em]
{\small\textbf{(g)}} & {\small\textbf{(h)}}
\end{tabular}

\vspace{0.15em}
\refstepcounter{figure}
\label{fig:qwen-chrome-showcase}
{\small\textbf{Figure \thefigure:} A representative Qwen for Chrome trajectory.
(a) A user requests the cheapest ECS server. (b) The agent proposes a purchase
plan and waits for confirmation. (c--d) It autonomously browses the site,
compares prices, and selects an option. (e) It configures the server. (f) Before
applying the configuration, it requests confirmation. (g) The specification
change completes successfully. (h) The agent pauses again before a risky disk
expansion operation.}
\end{center}

\clearpage
\section{Broader Computer-Use Related Work}
\label{app:broader-related-work}

\paragraph{Visual grounding and native agent models.}
The transition from structured web agents to general computer-use agents has
required models to understand dense, high-resolution interfaces and ground
language directly to screen coordinates. SeeClick, UGround, and OS-ATLAS study
this grounding problem across web, desktop, and mobile interfaces
~\citep{cheng2024seeclick,gou2024uground,wu2024osatlas}; CogAgent and Ferret-UI
develop vision-language models specialized for GUI and screen understanding
~\citep{hong2023cogagent,you2024ferretui}. More recent native agents integrate
grounding with planning, reasoning, and action prediction. Representative
systems include AutoGLM, Aguvis, UI-TARS, UI-TARS-2, and OpenCUA
~\citep{liu2024autoglm,xu2024aguvis,qin2025uitars,wang2025uitars2,
wang2025opencua,bai2025qwen3vl}. Qwen-CUA belongs to this latter family, with a
screenshot-only observation interface and native keyboard-and-mouse control.

\paragraph{Training data, tasks, and environments.}
Computer-use supervision has been collected from human demonstrations,
tutorial-guided replay, unlabeled interaction videos, and synthesized grounding
examples~\citep{wang2025opencua,xu2025agenttrek,lu2025videoagenttrek,
xie2025jedi}. A complementary line of work constructs executable environments
and verifiable tasks for online learning. ZeroGUI and related exploration-based
systems use learned evaluators to broaden task coverage, while GUI-GENESIS,
InfiniteWeb, and AutoWebWorld synthesize controllable web environments with
machine-checkable outcomes~\citep{yang2025zerogui,cao2026guigenesis,
zhang2026infiniteweb,wu2026autowebworld}. Gym-Anything broadens application
coverage, and CUA-Gym jointly scales reusable environments, task states, and
executable reward functions; SCALECUA further combines verifiable task synthesis
with capability-aware sampling for online training
~\citep{aggarwal2026gymanything,wang2026cuagym,lv2026scalecua}.

\paragraph{Rollout infrastructure.}
Large-scale agent learning also requires environment orchestration to be
decoupled from GPU-side optimization. NanoRollout exposes heterogeneous agent
harnesses and environment backends through a shared rollout service used by
evaluation, trajectory distillation, and reinforcement-learning clients
~\citep{wang2026nanorollout}. This systems abstraction complements cloud-scale
environment pools by allowing rollout workers and trainers to scale
independently.

\paragraph{Reinforcement learning and iterative improvement.}
GUI-R1 explores rule-based reinforcement learning for grounded GUI actions,
whereas UI-TARS-2 studies multi-turn reinforcement learning in interactive GUI
environments~\citep{xia2025guir1,wang2025uitars2}. ZeroGUI highlights the
importance of reward quality for stable online optimization
~\citep{yang2025zerogui}. EvoCUA-1.5 and SCALECUA extend this line with
multi-turn online optimization, adaptive data selection, and asynchronous or
segmented rollout processing~\citep{huang2026evocua15,lv2026scalecua}.
SEAgent and EvoCUA instead emphasize self-evolution:
an improved policy returns to the environment, generates new experience, and
learns again from selected trajectories~\citep{sun2025seagent,xue2026evocua}.
These directions motivate Qwen-CUA's combination of state-based rewards,
long-horizon trajectory optimization, and iterative refinement of supervised
data and RL tasks.

\paragraph{Computer-use evaluation.}
Web benchmarks progressed from offline demonstrations in Mind2Web to
interactive, reproducible environments such as WebArena, VisualWebArena, and
WorkArena~\citep{deng2023mind2web,zhou2023webarena,koh2024visualwebarena,
drouin2024workarena}. Desktop evaluation expanded through OmniACT, OSWorld,
OSWorld-Verified, Windows Agent Arena, and Computer Agent Arena
~\citep{kapoor2024omniact,xie2024osworld,bonatti2024windowsagentarena,
xlang2025osworldverified,wang2026computeragentarena}. Subsequent benchmarks
stress business processes,
long repetitive execution, personalized state, professional workflows, and
long-horizon cross-application work~\citep{wornow2024wonderbread,
wu2026osmarathon,jang2026mypcbench,sun2025scienceboard,yuan2026osworld2}.
Our evaluation complements these perspectives with broad capability,
efficiency, deployment, and hybrid-tool analyses.

\paragraph{Safety and robustness.}
As computer-use agents act on consequential state, evaluation must separate
task capability from safe execution. RedTeamCUA studies indirect prompt
injection in hybrid web--OS environments, OS-Harm covers deliberate misuse,
prompt injection, and model misbehavior, and RiOSWorld measures operational
risk in multimodal computer use~\citep{liao2025redteamcua,kuntz2025osharm,
yang2025riosworld}. These benchmarks motivate explicit confirmation before
consequential actions and reporting task success jointly with attack success or
other risk measures.

\end{document}